\documentclass[nohyperref]{article}

\usepackage{microtype}
\usepackage{graphicx}
\usepackage{subfigure}
\usepackage{booktabs} 
\usepackage{hyperref}

\usepackage[accepted]{icml2023}

\usepackage{amsmath,bm}
\usepackage{amssymb}
\usepackage{mathtools}
\usepackage{amsthm}

\usepackage[capitalize,noabbrev]{cleveref}

\usepackage{chngcntr}
\usepackage{bbm}

\newcommand{\name}{\textit{MyoDex}}

\icmltitlerunning{\name{}: A Generalizable Prior for Dexterous Manipulation}

\begin{document}
\setlength{\abovedisplayskip}{3pt}
\setlength{\belowdisplayskip}{3pt}

\twocolumn[
\icmltitle{\name{}: A Generalizable Prior for Dexterous Manipulation}

\begin{icmlauthorlist}
    \icmlauthor{Vittorio Caggiano}{meta}
    \icmlauthor{Sudeep Dasari}{cmu}
    \icmlauthor{Vikash Kumar}{meta,cmu}
\end{icmlauthorlist}

\icmlaffiliation{meta}{FAIR, Meta AI}
\icmlaffiliation{cmu}{CMU}

\icmlcorrespondingauthor{Vittorio Caggiano}{caggiano@fb.com}

\icmlkeywords{Musculoskeletal, Machine Learning, human dexterity}

\vskip 0.3in
]



\printAffiliationsAndNotice{}  % leave blank if no need to mention equal contribution

% \makeatletter
% \let\@oldmaketitle\@maketitle% Store \@maketitle
% \renewcommand{\@maketitle}{\@oldmaketitle% Update \@maketitle to insert...
%   \includegraphics[width=\linewidth,height=4\baselineskip]
%     {example-image}\bigskip}% ... an image
% \makeatother

\begin{figure*}[htpb]
     \centering
     \includegraphics[width=.95\textwidth]{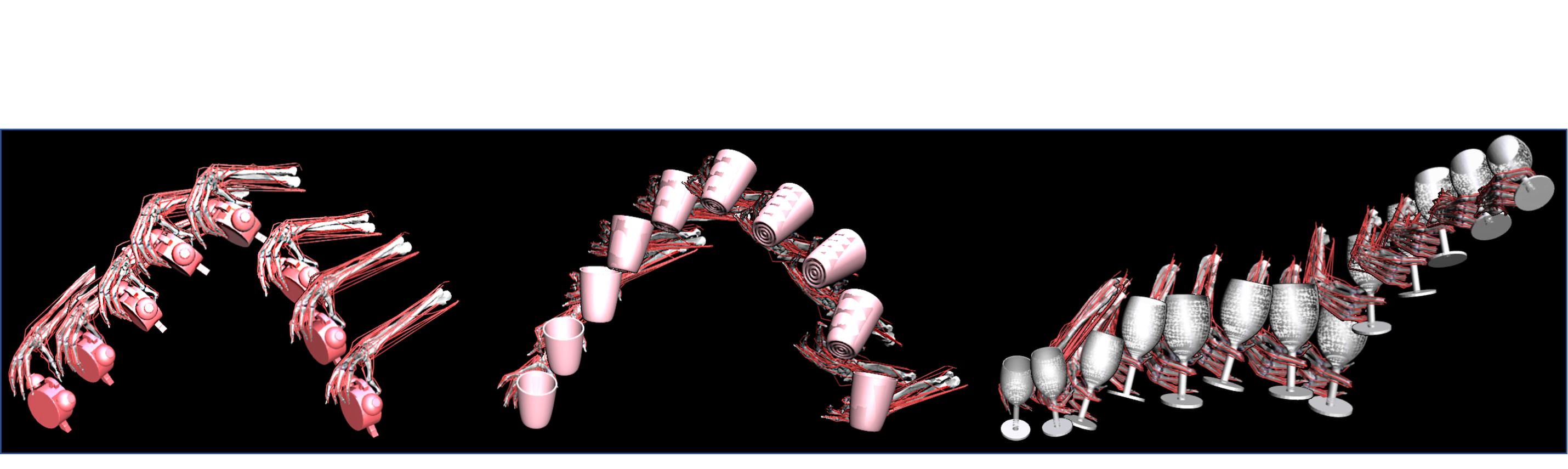}
     \caption{Contact rich manipulation behaviors acquired by \name{} with a physiological \textit{MyoHand}
     } \label{fig:Frames_Sequence}
\end{figure*}

\begin{abstract}

Human dexterity is a hallmark of motor control. Our hands can rapidly synthesize new behaviors despite the complexity (multi-articular and multi-joints, with 23 joints controlled by more than 40 muscles) of musculoskeletal sensory-motor circuits. In this work, we take inspiration from how human dexterity builds on a diversity of prior experiences, instead of being acquired through a single task.
Motivated by this observation, we set out to develop agents that can build upon their previous experience to quickly acquire new (previously unattainable) behaviors. Specifically, our approach leverages multi-task learning to implicitly capture task-agnostic behavioral priors (\name{}) for human-like dexterity, using a physiologically realistic human hand model -- MyoHand. We demonstrate \name{}’s effectiveness in few-shot generalization as well as positive transfer to a large repertoire of unseen dexterous manipulation tasks. Agents leveraging \name~ can solve approximately 3x more tasks, and 4x faster in comparison to a distillation baseline. 
While prior work has synthesized single musculoskeletal control behaviors, \name{} is the first \textit{generalizable} manipulation prior that catalyzes the learning of dexterous physiological control across a large variety of contact-rich behaviors. We also demonstrate the effectiveness of our paradigms beyond musculoskeletal control towards the acquisition of dexterity in 24 DoF Adroit Hand.

\textbf{Webpage}: \href{https://sites.google.com/view/myodex}{\color{blue}https://sites.google.com/view/myodex}

\end{abstract}

\section{Introduction}

Human dexterity (and its complexity) is a hallmark of intelligent behavior that set us apart from other primates species \cite{sobinov_neural_2021}. Human hands are complex and require the coordination of various muscles to impart effective manipulation abilities. New skills do not form by simple random exploration of all possible behaviors. Instead, human motor system relies on previous experience \cite{heldstab_when_2020} to build ``behavioral priors" that allow rapid skill learning \cite{yang_motor_2019, dominici_locomotor_2011, cheung_plasticity_2020}.

The key to learning such a prior might reside in the complexity of the actuation. Manipulation behaviors are incredibly sophisticated as they evolve in a high-dimensional search space (overactuated musculoskeletal system) populated with intermittent contact dynamics between the hands' degrees of freedom and the object. 
The human motor system deals with this complexity by activating different muscles as a shared unit. This phenomenon is known as a ``muscle synergy" \cite{bizzi_neural_2013}. Synergies allow the biological motor system -- via the modular organization of the movements in the spinal cord \cite{bizzi_neural_2013, caggiano_optogenetic_2016} -- to simplify the motor-control problem, solving tasks by building on a limited number of shared solutions \cite{davella_combinations_2003, davella_shared_2005}. Those shared synergies are suggested to be the fundamental building blocks for quickly learning new and more complex motor behaviors \cite{yang_motor_2019, dominici_locomotor_2011, cheung_plasticity_2020}. Is it possible for us to learn similar building blocks (i.e. a behavioral prior) for general dexterous manipulation on a simulated musculoskeletal hand?

In this work, we develop \name{}, a behavioral prior that allows agents to quickly build dynamic, dexterous, contact-rich manipulation behaviors with novel objects and a variety of unseen tasks -- e.g. drinking from a cup or playing with toys (see Figure \ref{fig:Frames_Sequence}). 
While we do not claim to have solved physiological dexterous manipulation, the manipulation abilities demonstrated here significantly advance the state of the art of the bio-mechanics and neuroscience fields. 
More specifically, our main contributions are: \textbf{(1)} We (for the first time) demonstrate control of a (simulated) musculoskeletal human hand to accomplish 57 different contact-rich skilled manipulation behaviors, despite the complexity (high degrees of freedom, third-order muscle dynamics, etc.). \textbf{(2)} We recover a task agnostic physiological behavioral prior -- \name \ -- that exhibits positive transfer while solving unseen out-of-domain tasks. Leveraging \name{}, we are able to solve 37 previously unsolved tasks. \textbf{(3)} Our ablation study reveals a tradeoff between the generality and specialization of the \name \ prior. The final system is configured to maximize \textit{generalization} and \textit{transfer} instead of zero-shot out-of-the-box performance. \textbf{(4)} We demonstrate the generality of our approach by applying it to learn behaviors in other high-dimensional systems, such as multi-finger robotic hands. We construct \textit{AdroitDex} (equivanet to \name{} for the \textit{AdroitHand} \cite{kumar2016manipulators}), which achieves 5x better sample efficiency over SOTA in the TCDM benchmark \cite{dasari_learning_2023}.

\section{Related Works}

Dexterous manipulations has been approached independently by the biomechanics field to study the synthesis of movements of the overactuated musculoskeletal system, and roboticists looking to develop, mostly via data-driven methods, skilled dexterous robots and a-priori representations for generalizable skill learning. Here, we discuss those approaches.

\textbf{Over-redundant biomechanic actuation.}
 Musculoskeletal models \cite{mcfarland_musculoskeletal_2021, lee_finger_2015, saul_benchmarking_2015,delp_opensim_2007, seth_opensim_2018} have been developed to simulate kinematic information of the muscles and physiological joints. Nevertheless, the intensive computational needs and restricted contact forces have prevented the study of complex hand-object interactions and otherwise limited the use mostly to optimization methods. Recently, a new hand and wrist model -- \textit{MyoHand} \cite{caggiano_myosuite_2022, wang_myosim_2022} -- overcomes some limitations of alternative biomechanical hand models: allows contact-rich interactions and it is suitable for computationally intensive data-driven explorations. Indeed, it has been shown that MyoHand can be trained to solve individual in-hand tasks on very simple geometries (ball, pen, cube) \cite{caggiano_myosuite_2022,MyoChallenge2022}. Here, we leveraged and extended the MyoHand model to perform hand-object manouvers on a large variaty of complex realistic objects.
 \begin{figure}[h]
 \vspace{-0.5cm}
     \centering
     \includegraphics[width=.45\textwidth]{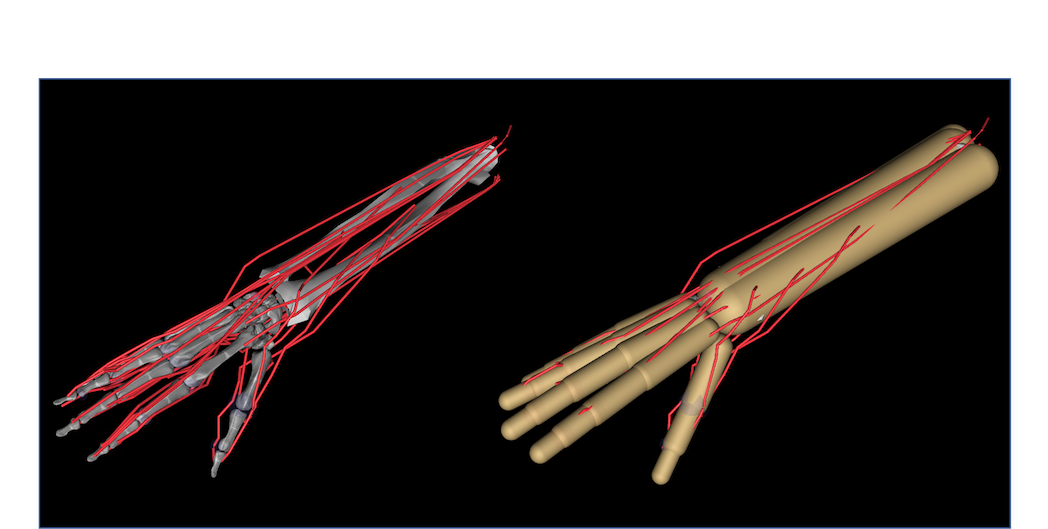}
     \caption{\textbf{\textit{MyoHand} - Musculoskeletal Hand model\cite{caggiano_myosuite_2022}.} On the left, rendering of the musculoskeletal structure illustrating bone -- in gray -- and muscle -- in red. On the right a skin-like surface for soft contacts is overlaid to the musculoskeletal model.}
     \label{fig:HandModel}
     \vspace{-0.5cm}
 \end{figure}

\textbf{Behavioral synthesis.} 
Data-driven approaches have consistently used Reinforcement Learning (RL) on joint-based control to solve complex dexterous manipulation in robotics \cite{rajeswaran_learning_2018, kumar_optimal_2016, nagabandi_deep_2019, chen_system_2021}.  In order to yield more naturalistic movements, different methods have leveraged motion capture data \cite{merel_learning_2017, merel_neural_2019, hasenclever_comic_2020}. By means of those approaches, it has been possible to learn complex movements and athletic skills such as high jumps \cite{yin_discovering_2021}, boxing and fencing \cite{won_control_2021} or playing basketball \cite{liu_learning_2018}. 

In contrast to joint-based control, in biomechanical models, machine learning has been applied to muscle actuators to control movements and produce more naturalistic behaviors. This is a fundamentally different problem than robotic control as the overactuated control space of biomechanical systems leads to ineffective explorations \cite{schumacher_dep-rl_2022}.  
Direct optimization \cite{wang_optimizing_2012, geijtenbeek_flexible_2013, al_borno_high-fidelity_2020, ruckert_learned_2013} and deep reinforcement learning \cite{jiang_synthesis_2019, joos_reinforcement_2020, schumacher_dep-rl_2022,ikkala_breathing_2022, caggiano_myosuite_2022, wang_myosim_2022, song_deep_2020, park_generative_2022} have been used
 to synthesize walking and running, reaching movements, in-hand manipulations, biped locomotion and other highly stylistic movements \cite{lee_dexterous_2018, lee_scalable_2019}. Nevertheless, complex dexterous hand-object manipulations beyond in-hand object rotation \cite{caggiano_myosuite_2022, berg_caggiano_kumar_2023} have not been accomplished so far.

\begin{figure*}[h!]
     \centering
     \vspace{3mm}
     \includegraphics[width=\textwidth]{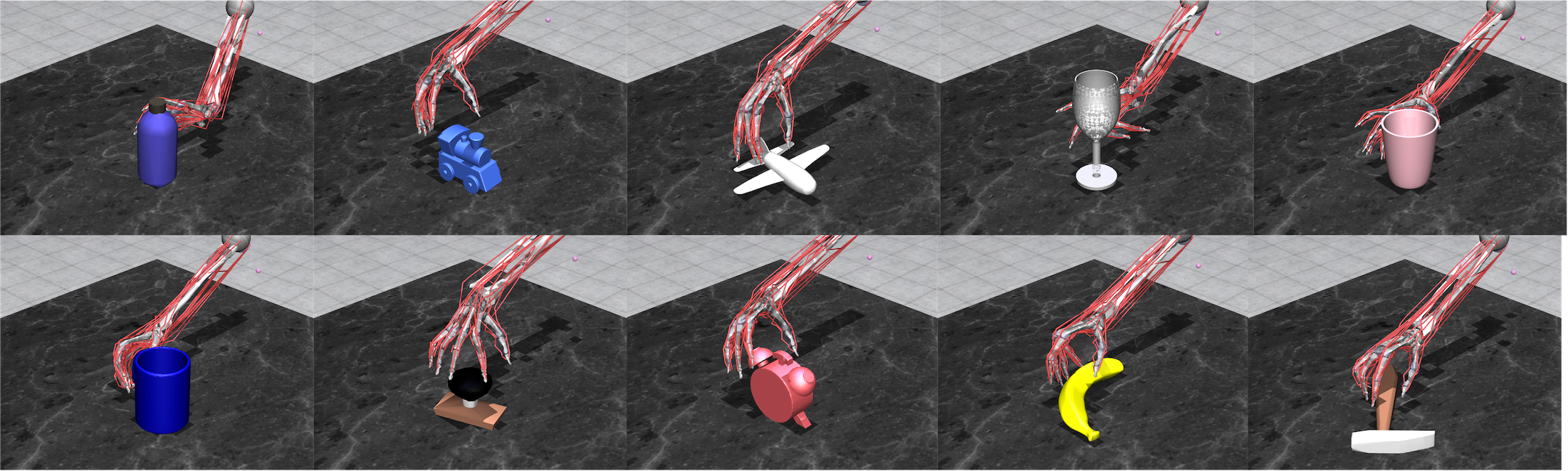}
     \caption{\textbf{Task setup and a subset of \textit{object-hand} pair from our task-set.} Every task setup consisted of a tabletop environment, an object, and the MyoHand. The MyoHand was shaped with a compatible posture and positioned near an object (i.e. pre-grasp posture).}
     \label{fig:Objects}
    %  \vspace{-7mm}
     \vspace{1mm}
\end{figure*}

\textbf{Manipulation priors.} 
Previous attempts have tried to solve complex tasks by building priors but this approach has been limited to games and robotics. 
The idea of efficiently representing and utilizing previously acquired skills has been explored in robotics by looking into features across different manipulation skills e.g. Associative Skill Memories \cite{pastor_towards_2012} and meta-level priors \cite{kroemer_learning_2016}. Another approach has been to extract movement primitives \cite{rueckert_extracting_2015} to identify a lower-dimensionality set of fundamental control variables that can be reused in a probabilistic framework to develop more robust movements. 

Multi-task learning, where a model is trained on multiple tasks simultaneously \cite{caruana_multitask_nodate}, has been also shown to improve the model's ability to extract features that generalize well \cite{zhang_regularization_2014, dai_instance-aware_2016, liu_end--end_2019}. Multi-task reinforcement learning (RL) has been used in robotics to propose representations-based methods for exploration and generalization in games and robotics \cite{goyal_infobot_2019, hausman_learning_2018}. However, training on multiple tasks can lead to negative transfer. As a consequence, performance on one task is negatively impacted by training on another task \cite{sun_adashare_2020}. Nevertheless, it has been argued that in (over)redundant control such as the physiological one, multi-task learning might facilitate learning of generalizable solutions \cite{caruana_multitask_nodate}. In this work, in addition to showing that nimble contact-rich manipulation using detailed physiological hand with musculoskeletal dynamics is possible, we present evidence that a generalizable physiological representation via Multi-task reinforcement learning -- \name \ -- can be acquired and used as priors to facilitate both learning and generalization across complex contact rich dexterous tasks.

\begin{figure*}[t!]
\setlength\intextsep{-0.4ex}
     \centering
     \includegraphics[width=\textwidth]{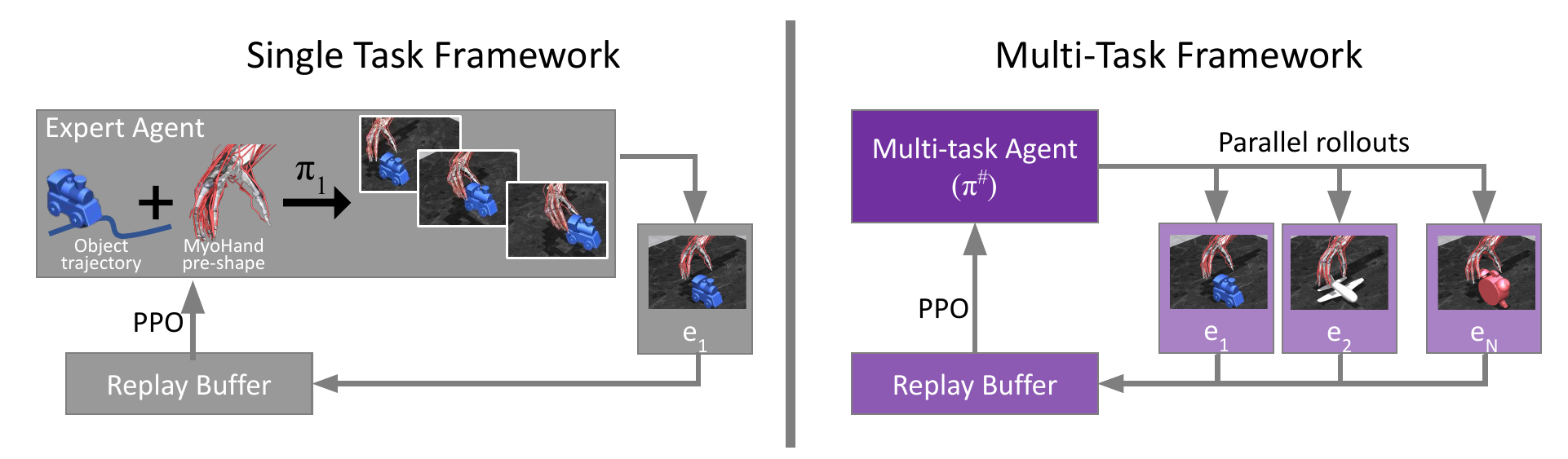}
     \caption{\textbf{Learning Frameworks.} Left - Single Task Framework: policies were obtained by training policies to solve the individual tasks. Right - Multi-task framework: A single policy (\name) was obtained by learning all tasks at once.}
     \label{fig:LearningParadigm}
     \vspace{1mm}
\end{figure*}

\section{Overactuated Physiological Dexterity}
Human hand dexterity builds on the fundamental characteristics of physiological actuation: muscles are multi-articular and multi-joints, the dynamics of the muscle are of the third order, muscles have pulling-only capabilities, and effectors have intermittent contact with objects. 
To further our understanding of physiological dexterity, we embed the same control challenges --  by controlling a physiologically accurate musculoskeletal model of the hand (see Sec. \ref{sec:MyoHand}) -- in complex manipulation tasks (see Sec. \ref{sec:task_design}).

\subsection{MyoHand: A Physiologically Accurate Hand Model} \label{sec:MyoHand}

In order to simulate a physiologically accurate hand model, a complex musculoskeletal hand model comprised of 29 bones, 23 joints, and 39 muscles-tendon units \cite{wang_myosim_2022} -- MyoHand model -- implemented in the MuJoCo physics simulator \cite{todorov_mujoco_2012} was used (see Figure \ref{fig:HandModel}). This hand model has previously been shown to exhibit a few dexterous \textit{in-hand} manipulation tasks \cite{caggiano_myosuite_2022}, which makes it a good candidate for our study seeking generalization in dexterous manipulation.

We extended the MyoHand model to include translations and rotations at the level of the shoulder. We limited the translation on the frontal (range between $[-0.07, \ 0.03]$) and longitudinal (range between $[-0.05, \ 0.05]$) axis to support the natural shoulder and wrist rotation (elbow is considered maximally extended i.e. a straight arm). For the rest of the paper we will refer to the whole system as \textit{MyoHand}.

\subsection{Dexterous Behaviors Studied}\label{sec:task_design}

In this study, we need a large variability of manipulations to explore the generality of our method against a wide range of solutions, hence it was important to include 1) objects with different shapes, and 2) complexity in terms of desired behaviors requiring simultaneous effective coordination of finger, wrist, as well as arm movements.

Our task set \textit{MyoDM} (inspired by TCDM benchmarks \cite{dasari_learning_2023}) is implemented in the MuJoCo physics engine \cite{todorov_mujoco_2012} and consists of 33 objects and 57 different behaviors. Every task setup (see Figure \ref{fig:Objects}) consists of a tabletop environment, an object from the ContactDB dataset \cite{brahmbhatt_contactdb_2019}, and the MyoHand. 

Dexterous manipulation is often posed as a problem of achieving the final desired configuration of an object. In addition to the final posture, in this study, we are also interested in capturing the detailed temporal aspect of the entire manipulation behavior. Tasks like drinking, playing, or cyclic movement like hammering, sweeping, etc., that are hard to capture simply as goal-reaching, can be handled by our formulation (Sec. \ref{sec:formulation}) and are well represented in the \textit{MyoDM}.

The tasks considered in \textit{MyoDM} entail a diverse variety of object manipulation (relocations+reorientations)  behaviors requiring synchronized coordination of arm, wrist, as well as in-hand movements to achieve the desired object behaviors involving simultaneous translation as well as rotation (average $\pm$ std, $28^{\circ} \pm 21^{\circ}$). The range of motions of the shoulder with fixed elbow alone is not sufficient to enable the entire range of desired object rotations without involving in-hand and wrist maneuvers. The angle between the palm and object ranges upwards of $20^{\circ}$ in our final acquired behaviors.
%average $\pm$ std, $8^{\circ} \pm 5^{\circ}$ with a maximum of $20^{\circ}$)
The wrist is one of the most complex joints to control because it is affected simultaneously by the balanced activation of more than 20 muscles whose activations also control finger movements. Careful maneuvering of objects within the hand requires simultaneous synchronization of numerous antagonistic finger muscle pairs, failing which leads to loss of object controllability; highlighting the complexities of controlling a physiological  musculoskeletal hand during these complex manipulations.

\section{Learning Controllers for Physiological Hands}
 \label{sec:formulation}
In this section, we discuss our approach to build agents that can learn contact-rich manipulation behaviors and generalize across tasks. 

\subsection{Problem formulation}  \label{sec:Reinforcement_Learning}

A manipulation task can be formulated as a Markov Decisions Process (MDP) \cite{sutton_reinforcement_2018} and solved via Reinforcement Learning (RL). In RL paradigms, the Markov decision process is defined as a tuple $\mathcal{M} = (\mathcal{S}, \mathcal{A}, \mathcal{T}, \mathcal{R}, \rho, \gamma)$, where  $\mathcal{S} \subseteq \mathbb{R}^n$ and $\mathcal{A} \subseteq \mathbb{R}^m$ represents the continuous state and action spaces respectively. The unknown transition dynamics are described by $s' \sim \mathcal{T}(\cdot|s,a)$. $\mathcal{R}: \mathcal{S} \rightarrow [0, R_{\max}]$, denotes the reward function, $\gamma \in [0, 1)$ denotes the discount factor, and and $\rho$ the initial state distribution. In RL, a policy is a mapping from states to a probability distribution over actions, i.e. ${\pi} : \mathcal{S} \rightarrow P(\mathcal{A})$, which is parameterized by $\theta$. The goal of the agent is to learn a policy $\pi_{\theta}(a|s) = argmax_{\theta}[J(\pi, \mathcal{M})]$, where $J = \max_{\theta} \mathbb{E}_{s_0 \sim \rho(s), a \sim \pi_{\theta}(a_t|s_t)}[\sum_t R(s_t, a_t)]$ %. 

\subsection{Learning Single-Task Controllers}
\textbf{Single task agents.}
 The single task agents are tasked with picking a series of actions ($[a_0,a_1,...,a_T]$), in response of the evolving states  ($[s_0,s_1,...,s_T]$) to achieve their corresponding  object's desired  behavior $\hat{X}_{object}=[\hat{x}_0, ... , \hat{x}_T ]$. 
 
 We adopt a standard RL algorithm \textit{PPO} \cite{schulman_proximal_2017} to acquire a goal-conditioned policy $\pi_{\theta}(a_t| s_t, \hat{X}_{object})$ as our single task agents. Details on state, actions, rewards, etc are provided in Section \ref{sec:State_Space}.  Owing to the third-order non-linear actuation dynamics and high dimensionality of the search space, direct optimization on $\mathcal{M}$ leads to no meaningful behaviors.

Pre-grasps implicitly incorporate information pertaining to the object and its associated affordance with respect to the desired task~\cite{jeannerod_neural_1988, santello_patterns_2002}. We leveraged \cite{dasari_learning_2023}'s approach of leveraging pregrasp towards dexterous manipulation with robotic (Adroit \cite{kumar2016manipulators}) hand  and extend it towards MyoHand.  The approach uses the hand-pose directly preceding the initiation of contact with an object i.e. a proxy to pre-grasp, to guide search in the high dimensional space in which dexterous behaviors evolve. This approach yeilds a set of single-task expert agents $\pi_i$ with $ i \in I$ where $I$ is the set of tasks (see Figure \ref{fig:LearningParadigm}-left).
 
\subsection{Framework for Multi-Task Physiological Learning} \label{section:PreGrasp_and_MT}

\textbf{Multi-task agent.}
Ideally, an agent would be able to solve multiple tasks using a goal-conditioning variable. Thus, we additionally train a single agent to solve a subset of tasks in parallel (see Figure \ref{fig:LearningParadigm}-right). This approach proceeds in a similar fashion as the single-task learner, but agent's experiences are sampled from the multiple tasks in \textit{parallel}. All other details of the agent $\pi_{\theta}^\#(a_t| s_t, \hat{X}_{object})$ (e.g. hyperparameters, algorithm, etc.) stay the same.

Similar to the single task agents, we encode manipulation behaviors in terms of goal-conditioned policies $\pi_{\theta}(a_t|s_t, \hat{X}_{object})$ and employ a standard implementation of the PPO \cite{schulman_proximal_2017} from Stable-Baselines \cite{raffin_stable-baselines3_2021} and pre-grasp informed formulation from \cite{dasari_learning_2023}'s to guide the search for our multi-task agents as well. See Section \ref{sec:PGDM} for details. The hyperparameters were kept the same for all tasks (see Appendix Table \ref{Table:Parameters}).

\section{Task Details}\label{sec:task_details}

Next, we provide details required to instantiate our \textit{MyoDM} task suite--

\textbf{State Space.}\label{sec:State_Space}
The state vector $s_t = \{\phi_t, \dot{\phi}_t, \psi_t, \dot{\psi}_t, \tau_t\}$ consisted of $\phi$ a 29-dimensional vector of 23 hand and 6 arm joints and velocity  $\dot{\phi}$, and object pose $\psi$ and velocity $\dot{\psi}$. 
In addition, positional encoding $\tau$ \cite{vaswani_attention_2017}, used to mark the current simulation timestep, was appended to the end of the state vector. This was needed for learning tasks with cyclic motions such as hammering.

\textbf{Action Space.}
The action space $a_t$ was a 45-dimensional vector that consists of continuous activations for 39 muscles of the wrist and fingers (to contract muscles), together with 3D translation (to allow for displacement in space), and 3D rotation of the shoulder (to allow for a natural range of arm movements).

\textbf{Reward Function.}
The manipulation tasks we consider involved approaching the object and manipulating it in free air after lifting it off a horizontal surface. The hand interacts with the object adjusting its positions and orientation ($X = [x_0, ..., x_T ]$) for a fixed time horizon. Similar to \cite{dasari_learning_2023}, this is translated into an optimization problem where we are searching for a policy that is conditioned on desired object trajectory $\hat{X} = [\hat{x}_0, ... , \hat{x}_T ]$ and optimized using the following reward function:

% \vspace{-2mm}
\begin{multline}
    % \vspace{-2mm}
    R(x_t,\hat{x}_t) := \lambda_1 exp\{-\alpha \| x_t^{(p)} - \hat{x}_t^{(p)} \|_2 - \\ \beta|\angle x_t^{(o)} - \hat{x}_t^{(o)}|   \} +  \lambda_2 \mathbbm{1}\{lifted\} -\lambda_3 \left \| \overline{m}_t  \right \|_2
    \label{Eq:Reward}
\end{multline}

where $ \angle $ is the quaternion angle between the two orientations, $\hat{x}_t^{(p)}$ is the desired object position, $\hat{x}_t^{(o)}$ is the desired object orientation, $\mathbbm{1}\{lifted\}$ encourages object lifting, and $\overline{m}_t$ the is overall muscle effort.

\textbf{Progress metrics.}\label{sec:metrics}
To effectively capture the temporal behaviors, we treat dexterous manipulation as a task of realizing desired object trajectories ($\hat{X}$). To capture temporal progress, similar to \cite{dasari_learning_2023}, we use three metrics to measure task performance. 
The \textit{success metric}, $ S(\hat{X}) $ reports the fraction of time steps where object error is below a $\epsilon = 1cm$ threshold. It is defined as: 
$ S(\hat{X})=\frac{1}{T} \sum_{t=0}^{T} \mathbbm{1}  {\left \| x_t^{(p)} - \hat{x}_t^{(p)}   \right \|_2 < \epsilon} $. \label{Eq:MetricSuccess}
The \textit{object error metric} $E(\hat{X})$ calculates the average Euclidean distance between the object’s center-of-mass position and the desired position from the desired trajectory:
$ E(\hat{X})=\frac{1}{T}\sum_{t=0}^{T} \left \| x_t^{(p)} - \hat{x}_t^{(p)}   \right \|_2 \label{Eq:MetricError}$.  In addition, we also used the \textit{object orientation metric}:  $ O(\hat{X})=\frac{1}{T}\angle_{t=0}^{T} ( x_t^{(o)} - \hat{x}_t^{(o)}   ) \label{Eq:OriError}$
\footnote{For interpretability, we often omit orientations because center-of-mass error and orientation error were highly correlated in practice i.e. Pearson-correlation $> 0.785$}.

\begin{figure}[h]
     \centering
     \vspace{3mm}
     \includegraphics[width=0.35\textwidth]{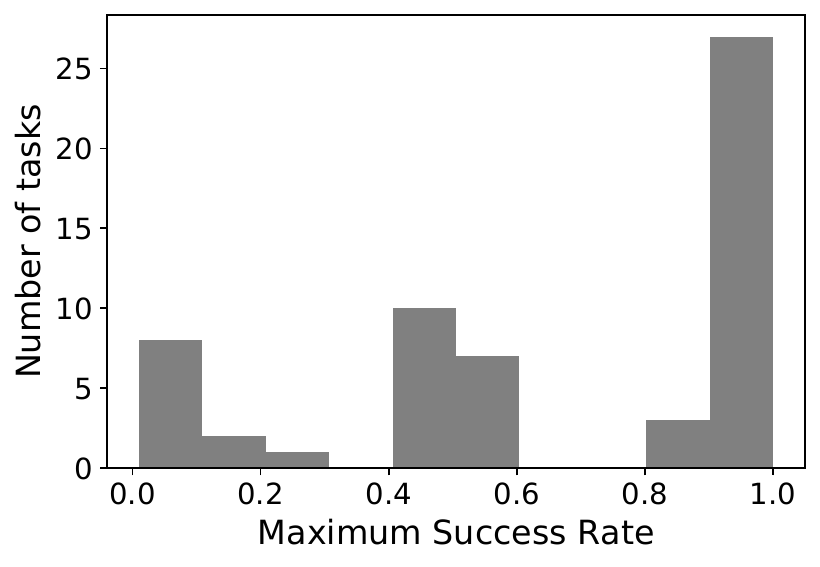}
     \caption{\textbf{Distribution of single task solutions.} Distribution of maximums success rate for single-task solutions on 57 different tasks. Only 32 out of 57 tasks i.e. $56\%$, were solved with a success rate above $80\%$. Training performed over $12.5k$ iterations.}
     \label{fig:DistributionSuccessRate_Experts}
\end{figure}

\section{Results}
\label{sec:result}

First, we study if we can solve the \textit{MyoDM} task set, one task at a time (see Sec.~\ref{sec:result_single_task}). Next, we illustrate that our \name \ representation can be used as \textit{a prior} for accelerating learning novel, out-of-domain tasks (see Sec.~\ref{sec:result_generalization}). Finally, we present a series of ablation studies to understand various design choices of our approach (see Sec.~\ref{sec:result_ablation}).

\subsection{Learning Expert Solutions for Single-Task Setting} \label{sec:result_single_task}
We begin by asking, is it possible to learn a series of complex dexterous manipulation behaviors (see Sec. \ref{sec:task_design}) using a MyoHand? Our single-task learning framework is applied to solve a set of 57 \textit{MyoDM} tasks independently, without any object or task-specific tuning (see Table \ref{Table:Parameters}). The resulting ``expert policies" were able to properly manipulate only a subset of those objects, while moving and rotating them to follow the target trajectory (see Figure \ref{fig:Frames_Sequence} for a sequence of snapshots). This was quantified by using 2 metrics (Sec. \ref{sec:metrics}): a Success Metric and an Error Metric. 
Our single-task framework achieves an average success rate of $66\%$ solving 32 out of 57 tasks (see Fig. \ref{fig:DistributionSuccessRate_Experts} and experts in Fig. \ref{fig:alltasks}) and an average (ecludean distance) error of $0.021$. We encourage readers to check our \href{https://sites.google.com/view/myodex}{project website} for videos and further qualitative analysis of the learned behaviors.

\begin{figure}[h!]
     \centering
     \includegraphics[width=.45\textwidth]{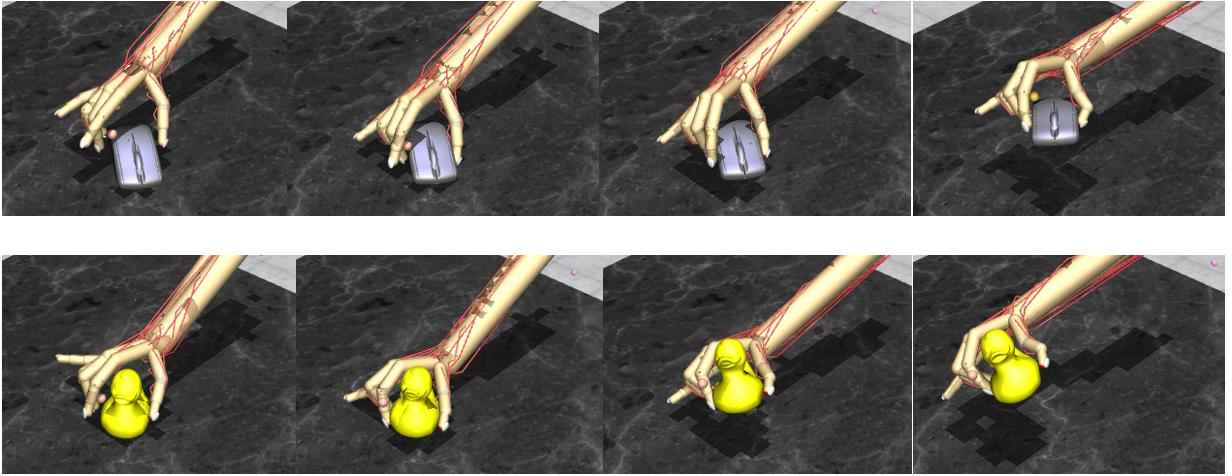}
    
     \caption{\textbf{Zero-shot generalization.} \name{} successfully initiated manipulations on new objects and trajectories. Hand rendering includes skin-like contact surfaces (see Fig. \ref{fig:HandModel})}
     \label{fig:Zero_Shot_Gen}
    
\end{figure}

\begin{figure*}[h!]
     \centering
     \includegraphics[width=.95\textwidth]{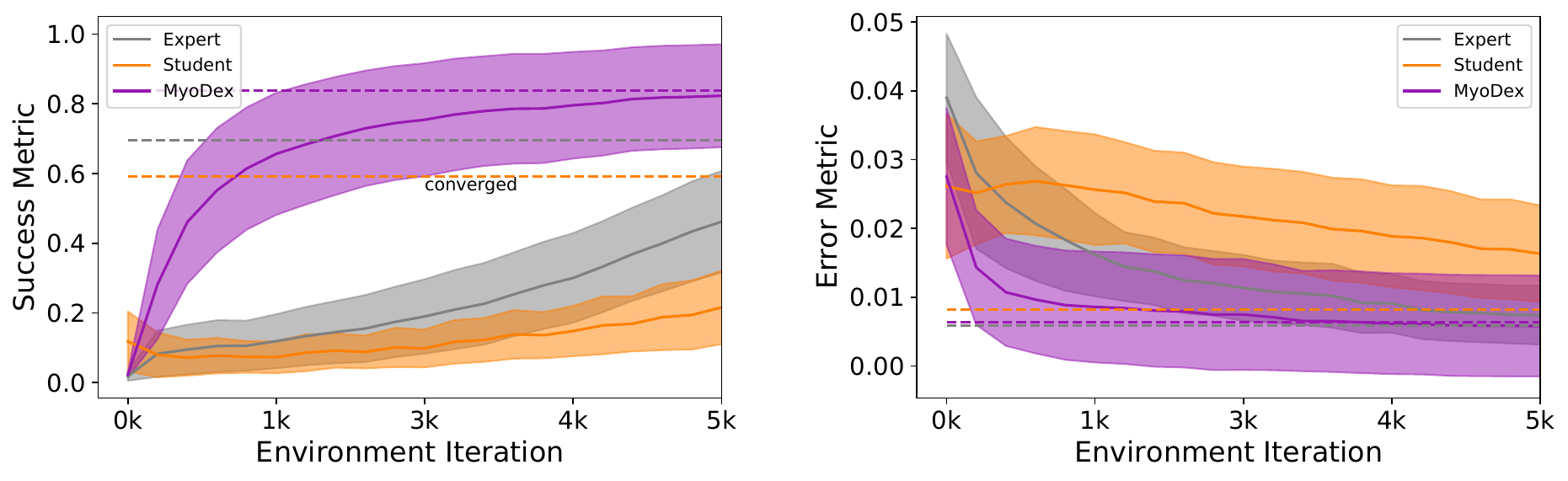}
     \caption{\textbf{Fine-tuning on 43 Out-of-domain tasks.} Metrics until $5k$ iterations of the fine tuning of 43 out-of-domain tasks. Convergence is assumed at $12.5k$ iterations.  Left - Success Metric. Right - Error Metric. Continuous lines show average and shaded areas the standard deviation of success and error metrics. The dashed line represents the value at convergence i.e. $12.5k$ iterations.}
     \label{fig:FineTuning_ood_curves}
     % \vspace{-7mm}
\end{figure*}

\subsection{Accelerating Out-of-Domain Learning via \name~} \label{sec:result_generalization}

The single-task framework was not able to solve all task in our task set, even individually which further establishes complexity of behavior acquisition with high DoF MyoHand and the difficulty of our \textit{MyoDM} task set. Furthermore, it creates controllers that can only function within a specific scenario/task. Next, we will demonstrate that by simultaneously training on multiple tasks during the reinforcement learning loop we can achieve a \name \ prior that can overcome single-task limitations. \name \ is a prior that can be fine-tuned to solve a larger amount of tasks.  In addition, a single multi-task policy based on training \name \ at convergence can solve multiple tasks.

For building the \name \ prior, we consider a subset of 14 \textit{MyoDM} tasks with a large variability of object and movements (see Sec. \ref{sec:task-choice} for the effects of task choice) 
and we trained one policy to solve all the set of tasks at once. We stopped the training at $12.5k$ iterations (at the beginning of the error plateau -- see Figure \ref{fig:Multi-Task_training_curve_zoom_in}). At this iteration, we tested potential \textit{zero-shot} generalization capabilities with the MyoHand positioned near a novel object with a compatible posture and conditioned on a new task trajectory. While the policy was not able to zero-shot solve these new tasks (success rate $\leq 10\%$), we do observe (see Fig.~\ref{fig:Zero_Shot_Gen}) that the hand can succesfully grasp and lift the unseen objects. This leads us to believe that the \name \ representation can be used as a \textit{prior} for accelerating transfer learning.

However, this is not the only way to accomplish a general multi-task representation. An established baseline is a student-teacher distillation (see Sec. \ref{section:imitation_learning}), which trains a single student policy to imitate the 14 expert policies (from prior experiments) via behavior cloning.

We fine-tune both the \name \ and the student policy on the remaining  
out-of-domain set of 43 \textit{MyoDM} tasks (using single-task RL) for additional iterations. Figure~\ref{fig:FineTuning_ood_curves} presents learning curves for the fine-tuned models based on \name{}, fine-tuned student baselines, and trained (from scratch) single-task expert policies in terms of success rate and errors, respectively. Note how the \name \ based policy is able to learn the tasks significantly faster than either the baseline or the single-task policies. Among the solved out-of-domain tasks, \name{} based policies were about $4$x faster than student based policy ($1.9k$ vs $7.7k$), and approximately $3$x fastern than single-task expert policy ($1.9k$ vs $5.7k$, Table \ref{table:FineTuning_summary}). Additionally, it achieves a \textit{higher overall task performance in comparision to the single-task experts}, which plateau at a significantly lower success rate, likely due to exploration challenges. Table~\ref{table:FineTuning_summary} shows this trend in extra detail. The \name \ representation allows to solve more tasks ($37$ vs $22$, see Table \ref{table:FineTuning_summary} and Table \ref{table:FineTuning}) and achieve a higher overall success rate ($0.89$ vs $0.69$) than the single-task expert, which in turn outperforms the student baseline. This leads us to conclude that the \name~ representation can act as a generalizable prior for learning dexterous manipulation policies on a musculoskeletal MyoHand. It is both able substantially accelerate learning new tasks, and indeed leads to a \textit{stronger} transfer to new tasks.

\begin{table}[h]

    \centering
    \resizebox{\linewidth}{!}{
        \begin{tabular}{|l|c|c|c|}
        \hline
        \textbf{Based on} &  \textbf{Solved} &  \textbf{Success} &\textbf{Iter. to solve}  \\
        \hline
        Expert	&  $51\%$ (22/43) &  $0.69 \pm 0.30$  & $5.7k\pm1.5k$ \\
        Student	& $30\%$ (13/43) 	& $0.54 \pm 0.35$   & $7.7k \pm 1.9k$\\
        \textbf{\name{}} & $\boldsymbol{86\%}$ \textbf{(37/43)}	& $\boldsymbol{0.89 \pm 0.25}$	& $\boldsymbol{1.9k \pm 2.1k}$ \\
        \hline    
    \end{tabular}
    }
    \caption{\textbf{\name{} transfer statistics on unseen (43) tasks} -- \textit{Solved} indicates the percentage (ratio) of solved tasks (success $\geq80\%$). 
    \textit{Success} indicates the success metric stats on all 43 tasks at $12.5k$ iterations.
    \textit{Iter. to solve} indicates the stats on min iterations required by the solved task to achieve $\geq80\%$ success. Values are expressed as average $\pm$ std.}
    \vspace{-.5cm}
    
    \label{table:FineTuning_summary}
\end{table}

\subsection{Multi-Task Learning with \name~} \label{sec:result_multitask}
Additionally, \name \ can also be used to recover one single policy that can solve multiple tasks. We compared the results of the \name \ training at convergence against the student policy (from the distillation of experts) on the same set of 14 tasks. See a summary of the results Figure~\ref{fig:MetricsDistilled_in-domain}. The converged \name \ based policy's success rate improves by $>2$x over the student policy. 
We present an explanation in Section \ref{sec:Discussion_synergies} of why distilling from experts that have acquired incompatible behaviors in an over-redundant musculoskeletal system fails at learning multi-task policies. Indeed, expert policies found a local solution that does not help to learn other tasks e.g. experts used as a-priori do not help to fine-tune other tasks (see Fig. \ref{Fig:FineTuning_Experts}). In contrast, our multi-task framework avoids this pitfall, since it simultaneously learns one policy without any implicit bias, and can reach similar levels as reached by individual experts in isolation.

\subsection{\name~Ablation Study} \label{sec:result_ablation}
The previous set of experiments demonstrated that \name \ contains generalizable priors for dexterous manipulation. The following ablation study investigates how changing the number of pre-training iterations as well as the number of tasks during pre-training affect the \name's capabilities.

\subsubsection{Effects of iterations on the \name \ representation}\label{sec:FT_checkpoint}
\begin{figure}[h!]
     \centering
     \includegraphics[width=.4\textwidth]{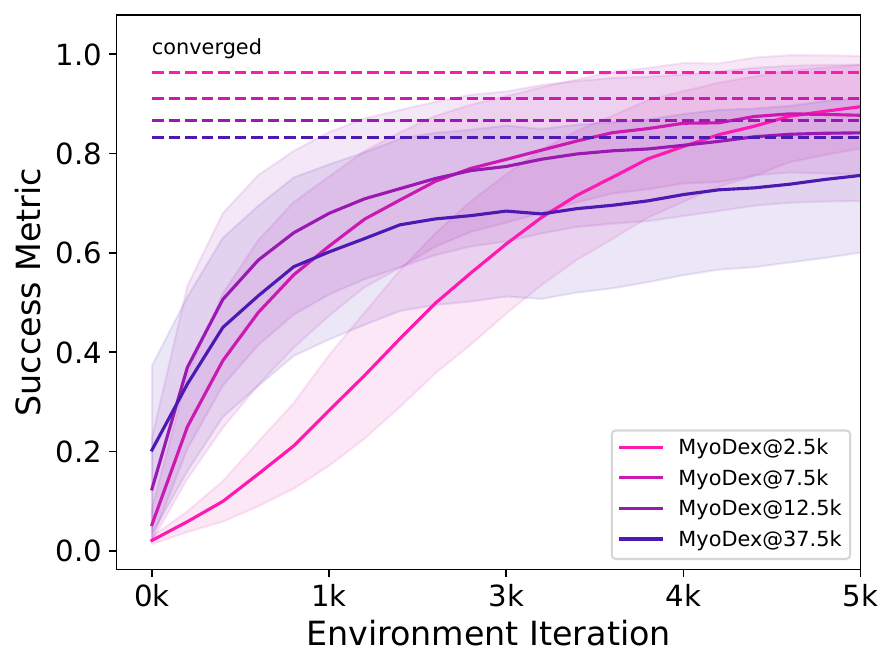}
     \caption{\textbf{Fine-tuning from representations obtained at different iterations.} Fine-tuning from representations obtained earlier i.e. $2.5k$ and $7.5k$ iterations, \name~ i.e. $12.5k$ iterations, and later i.e. $37.5k$ iterations. Earlier representations show no few-shot generalization but better coverage with 56 out of 57 tasks solved, while later representations show few-shot generalizations but have less coverage with 35 out of 57 tasks solved. The continuous line represents the average and the shaded area is the standard deviation of the success metrics. The dashed line represents the value at convergence i.e. 12.5k iterations.}
     \label{fig:FT_checkpoints}
\end{figure}

In our experiment, the multi-task policy at $12.5k$ iterations is defined as the \name \ prior. At this number of iterations, the policy was able to achieve $\sim35\%$ success rate (see Fig. \ref{fig:Multi-Task_training_curve_zoom_in}). This solution provided both few-shot learning (task solved within the first environment iteration) 
most of the \textit{MyoDM} set of 57 tasks. Here, in order to probe the sensitivity of \name \ prior  to the stage of learning at which the representation is extracted, we compared ~\name \ against representations obtained earlier i.e. $2.5k$ and $7.5k$, and one later i.e. $37.5k$ stages of learning. Figure \ref{fig:FT_checkpoints} shows the results on the fine-tuning of all the 57 tasks for the 4 different representations. Early representations are slower but, with enough iterations, they are able to solve almost all tasks ($98\%$ (56 / 57) and $91\%$ (52 / 57) respectively for the representations at $2.5k$ and $7.5k$). Conversely, later representations, show few-shot learning (10 tasks) but they are able to learn only a reduced amount of tasks ($61\%$ (35 / 57)). Hence, \name~ trained at $12.5k$ iterations strikes a balance, facilitating fast initial learning (including few-shots) while being general enough to support a diverse collection of out-of-domain tasks (see Figure \ref{fig:FT_checkpoints}).

Another way to look at the effect of learning and generalizable solutions over the iterations is to look to muscle synergies as they express the amount of muscle co-contraction shared across tasks. In our study, we utilized the concept of Variance Accounted For (VAF, see Sec. \ref{section:muscle_synergies}) to quantify the number of synergies 
needed to reconstruct the needed muscle activations to solve the task. 
Higher VAF achieved with fewer muscle synergies indicates that it is possible to use fewer combinations of muscle co-contractions to generate the needed muscle activations. Our findings indicate that early on in the training process (i.e., around 2.5k iterations, see Figure \ref{fig:Synergies_iterations}), a substantial number of synergies (more than 12) is needed to achieve a high level of signal reconstruction. This suggests that while the policy is capable of discovering some solutions in the muscle space, synergies are not able to cover all the variability of the signal. Indeed, this representation helps to overcome some local minima hence it is particularly well-suited for facilitating transfer to new tasks.

Around 12.5k iterations, we observed a peak in the capacity of fewer synergies to account for most of the signal (see Figure \ref{fig:Synergies_iterations}). At this point we have identified solutions in the muscle space that are highly reusable across multiple tasks.

However, at 37.5k iterations, we found that a greater number of synergies were required to explain most of the original signals. This indicates that specialized co-contractions are emerging to address specific tasks demands. While these synergies are effective at solving similar tasks with few or zero shots, their specialization may limit their ability to tackle dissimilar tasks.

Overall, our results suggest that our representation of synergies is a powerful tool for facilitating transfer learning, especially in the early stages of training when more generalized solutions are required. As training progresses, the emergence of specialized co-contractions enables efficient learning and transfer to similar 
 tasks. Still, with even more training, specialized solutions are developed. 

\subsubsection{Effect of the number of environments on \name~ training}
In the above experiment, we showed the \name \ representation based on 14 environments.  An analysis showing the effect of multi-task learning on environment diversity illustrates that the use of 14 environments represented a balance  between trainign the multi-task policy effectively and transfer/generalization ability it possses. We compared \name \ trained on 6, 14, and 18 environments at $12.5k$ iterations and tested on a set of 39 new environments. \name \ based on 6 and 18 environments leads to lower performance with respect to 14 environments both in terms of success rate and the number of solved environments (see Table \ref{table:FineTuning_different_priors}). 

\begin{table}[h]

    \centering
        \begin{tabular}{|l|c|c|}
        \hline
        \textbf{Based on} &  \textbf{Success} &  \textbf{Solved} \\
        \hline 
        \name{}6	 &  $0.78 \pm 0.32$ & $72\% \ (28/39)$ \\
        \textbf{\name{}14}  & $\boldsymbol{0.92 \pm 0.21}$ & $\boldsymbol{95\% \ (37/39)}$ \\
        \name{}18  & $0.91 \pm 0.2$  & $87\% \ (34/39)$ \\
        \hline    
    \end{tabular}
    \caption{\textbf{Fine-tuning statistics based on different \name \ priors.} \name \ trained with different environments as priors and fine-tuned on 39 environments. Results reported in terms of average and standard deviation of success and percentage of solved tasks i.e. $\geq 80\%$.  }
    \label{table:FineTuning_different_priors}
\end{table}

\begin{figure}[h!]
     \centering     
     \includegraphics[width=0.4\textwidth]{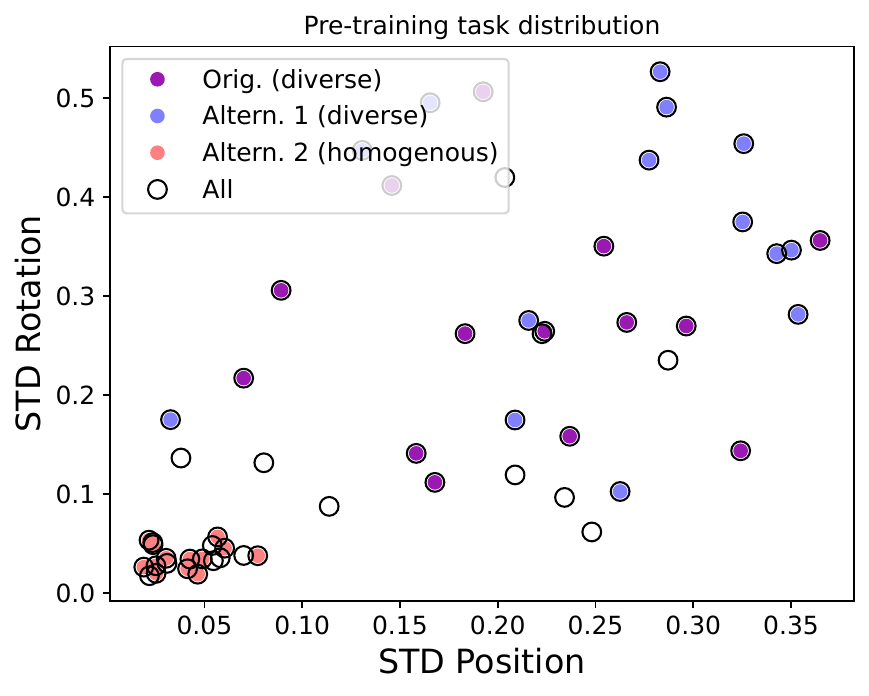}
     \caption{\textbf{Pre-Training task distribution. }   The distributions of our task collection in terms of its variability (standard deviation - STD). 
    Represented on each axes are the STD of the absolute positional (X-Axis) and rotational (Y-axis) displacements from the respective initial object poses in the desired object trajectories in our task set.
    In circle are all the 57 tasks involved in our study
    In pink [Orig.(diverse)] are the original tasks used for training MyoDex.
    In blue [Altern.1(diverse)] is a new task set we use for training an alternate instance of MyoDex prior used in ablation studies.}
     \label{fig:Task_Complexity}
\end{figure}
\subsubsection{How Training Tasks Affect \name{}}
\label{sec:task-choice}
\begin{figure}
    \centering
    \includegraphics[width=0.4\textwidth]{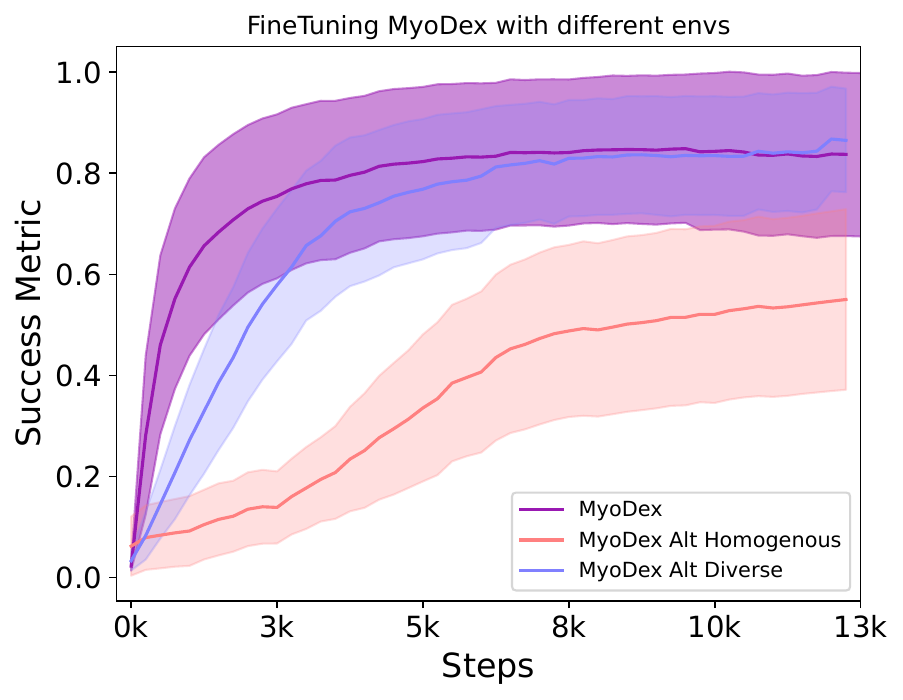}
    \caption{\textbf{Effect of pre-training task distribution on \name{} performance. } \textit{MyoDex Alt Diverse} (trained on tasks of similar diversity – in blue) is able to better match the original \name{} performance in comparision to \textit{MyoDex Alt Homogenous} (trained on homogenous tasks collection). }
    \label{fig:Success_MyoDexAlternatives}
\end{figure}

The choice of objects and tasks to train \name{} can significantly impact the effectiveness of the representation. We study the effect of pre-training task distribution on the effectiveness of MyoDex priors. We selected two new task sets. First, a \textit{diverse} tasks collection -- \textit{MyoDex Alt Diverse} (Figure \ref{fig:Task_Complexity} in blue) with the same similar attributes of the original dataset (in pink). Second, a \textit{homogenous} task collection -- \textit{MyoDex Alt Homogenous} (Figure \ref{fig:Task_Complexity} in red) -- with tasks with little motion variance (e.g. mostly lifting). We found that \textit{MyoDex Alt Diverse} – trained on the alternative diverse tasks – was able to improve performance over time, while \textit{MyoDex Alt Homogenous} – trained on the alternative homogenous tasks – had its performance plateau early on during training (see Figure \ref{fig:TaskDistr_PreTraining}). 
Indeed, when used for transfer on new tasks, \textit{MyoDex Alt Diverse} is able to match the original \name{} performance, while \textit{MyoDex Alt Homogenous} does not (see Figure \ref{fig:Success_MyoDexAlternatives}). This shows that the variety of manipulation/tasks in the pretraining is fundamental to achieve high performance in a larger set of downstream tasks and \name{} is not sensitive to the specific choice of tasks.

\subsection{Extension to other high dimensional system}\label{section:AdroitDex}
\begin{figure}[h!]
     \centering
     \includegraphics[width=.4\textwidth]{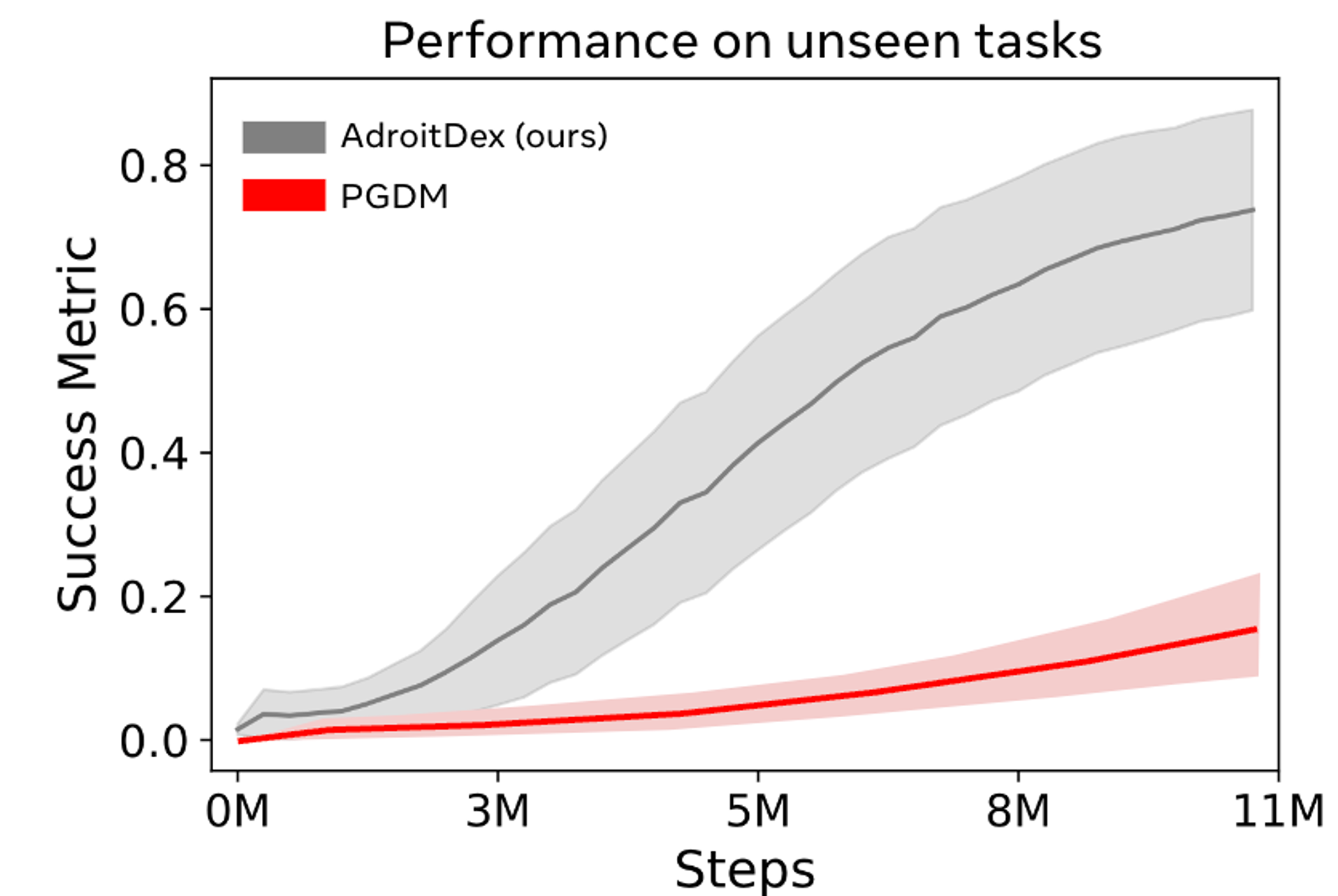}
     \caption{ \textbf{Fine-tuning a generalizable representation on Adroit subtasks: \textit{AdroitDex}.} A general representation of manipulation on the same 14 tasks used for trainign \name{} was finetuned on 34 unseen tasks on the TCDM benchmark \cite{dasari_learning_2023}. Curves shows average (continuous) and std (shaded area). \textit{AdroitDex} beats previously reported SOTA on TCDM benchmarks while being 5x more sample efficient.}
     \label{fig:AdroitDex}
\end{figure}

To further investigate the applicability of our approach to other high dimensional systems, we set out to build a generalizable representation for the robotic \textit{Adroit Hand }\cite{rajeswaran_learning_2018} commonly studied in robot learning. Adroit is a 24 degree-of-freedom (DoF) modified shadow hand with 4 extra DoF at the distal joint. Following the approach behind \name{}, a general representation of manipulation prior - \textit{AdroitDex} - was obtained. We use the same 14 tasks that we used for training \name{}. In the Figure \ref{fig:AdroitDex} we show the performance of \textit{AdroitDex} on 34 unseen tasks on the TCDM benchmark~\cite{dasari_learning_2023}. \textit{AdroitDex} achieves a success rate of $74.5\%$ in about 10M iteration steps, which is approvimately 5x faster than the PGDM baseline \cite{dasari_learning_2023}, which needed 50M iteration steps to achieve the same result (see Figure \ref{fig:AdroitDex}).

\section{Conclusion}
\label{sec:conclusion}
In this manuscript, we learn skilled dexterous manipulation of complex objects on a musculoskeletal model of the human hand. In addition, by means of joint multi-task learning, we showed that it is possible to extract generalizable representations (\name) which allow faster fine-tuning on out-of-domain tasks and multi-task solutions. Ultimately, this study provides strong bases for how physiologically realistic hand manipulations can be obtained by pure exploration via Reinforcement Learning i.e. without the need for motion capture data to imitate specific behavior. 

\section{Discussion on the role of Synergies}\label{sec:Discussion_synergies}
Why does \name \ help the overactuated musculoskeletal system to solve multiple tasks? If we look at the coordination of muscle activations -- muscle synergies (see Appendix \ref{section:muscle_synergies}) -- we notice that \name \, shows a larger number of similar activations (see Figure \ref{Fig:CosSimilarity}) vs experts/distilled policies. This is because the expert solutions find one mode/solution to solve a task that does not incorporate information from other tasks. Naiive distillation propogates this effect to the student policy. In contrast, \name \ learns to coordinate muscle contraction. Indeed, fewer muscle coordination/synergies seem to explain most of the behavior (see Figure \ref{fig:Synergies_iterations}, at 12.5K iterations). All in all, those observations are in line with the neuroscience literature where muscle synergies have been suggested as the physiological substrate to obtain faster and more effective skill transfer \cite{yang_motor_2019,cheung_plasticity_2020,dominici_locomotor_2011,berger_differences_2013}.

\section{Limitations and Future work}
\label{sec:future_work}
While we demonstrated that \name{} can produce realistic behavior without human data, one important limitation is understanding and matching the results with physiological data. Indeed, our exploration method via RL produced only one of the very high dimensional combinations of possible ways that a human hand could hypothetically grab and manipulate an object. For example, there are several valid ways to hold a cup e.g. by using the thumb and one or multiple fingers. Although our investigation points us in the right direction regarding the physiological feasibility of the result, these findings have yet to be properly validated with clinical data and user studies. Future works will need to consider the ability to synthesize new motor behaviors while simultaneously providing muscle validation.

%============================================================



\bibliography{References}

\begin{thebibliography}{65}
\providecommand{\natexlab}[1]{#1}
\providecommand{\url}[1]{\texttt{#1}}
\expandafter\ifx\csname urlstyle\endcsname\relax
  \providecommand{\doi}[1]{doi: #1}\else
  \providecommand{\doi}{doi: \begingroup \urlstyle{rm}\Url}\fi

\bibitem[Al~Borno et~al.(2020)Al~Borno, Vyas, Shenoy, and
  Delp]{al_borno_high-fidelity_2020}
Al~Borno, M., Vyas, S., Shenoy, K.~V., and Delp, S.~L.
\newblock High-fidelity musculoskeletal modeling reveals that motor planning
  variability contributes to the speed-accuracy tradeoff.
\newblock \emph{{eLife}}, 9:\penalty0 e57021, 2020.
\newblock ISSN 2050-084X.
\newblock \doi{10.7554/eLife.57021}.
\newblock URL \url{https://doi.org/10.7554/eLife.57021}.

\bibitem[Berg et~al.(2023)Berg, Caggiano, and Kumar]{berg_caggiano_kumar_2023}
Berg, C., Caggiano, V., and Kumar, V.
\newblock Sar: Generalization of dexterity via synergistic action
  representation.
\newblock \url{ https://sites.google.com/view/sar-rl/home }, 2023.

\bibitem[Berger et~al.(2013)Berger, Gentner, Edmunds, Pai, and
  d'Avella]{berger_differences_2013}
Berger, D.~J., Gentner, R., Edmunds, T., Pai, D.~K., and d'Avella, A.
\newblock Differences in adaptation rates after virtual surgeries provide
  direct evidence for modularity.
\newblock \emph{Journal of Neuroscience}, 33\penalty0 (30):\penalty0
  12384--12394, 2013.
\newblock ISSN 0270-6474, 1529-2401.
\newblock \doi{10.1523/JNEUROSCI.0122-13.2013}.
\newblock URL \url{https://www.jneurosci.org/content/33/30/12384}.
\newblock Publisher: Society for Neuroscience Section: Articles.

\bibitem[Bizzi \& Cheung(2013)Bizzi and Cheung]{bizzi_neural_2013}
Bizzi, E. and Cheung, V.~C.
\newblock The neural origin of muscle synergies.
\newblock \emph{Frontiers in Computational Neuroscience}, 7, 2013.
\newblock ISSN 1662-5188.
\newblock \doi{10.3389/fncom.2013.00051}.
\newblock URL
  \url{https://www.frontiersin.org/article/10.3389/fncom.2013.00051}.

\bibitem[Brahmbhatt et~al.(2019)Brahmbhatt, Ham, Kemp, and
  Hays]{brahmbhatt_contactdb_2019}
Brahmbhatt, S., Ham, C., Kemp, C.~C., and Hays, J.
\newblock {ContactDB}: Analyzing and predicting grasp contact via thermal
  imaging.
\newblock \emph{2019 {IEEE}/{CVF} Conference on Computer Vision and Pattern
  Recognition ({CVPR})}, pp.\  8701--8711, 2019.

\bibitem[Caggiano et~al.(2016)Caggiano, Cheung, and
  Bizzi]{caggiano_optogenetic_2016}
Caggiano, V., Cheung, V. C.~K., and Bizzi, E.
\newblock An optogenetic demonstration of motor modularity in the mammalian
  spinal cord.
\newblock \emph{Scientific Reports}, 6\penalty0 (1):\penalty0 35185, 2016.
\newblock ISSN 2045-2322.
\newblock \doi{10.1038/srep35185}.
\newblock URL \url{https://doi.org/10.1038/srep35185}.

\bibitem[Caggiano et~al.(2022{\natexlab{a}})Caggiano, Wang, Durandau, Sartori,
  and Kumar]{caggiano_myosuite_2022}
Caggiano, V., Wang, H., Durandau, G., Sartori, M., and Kumar, V.
\newblock {MyoSuite} – a contact-rich simulation suite for musculoskeletal
  motor control, 2022{\natexlab{a}}.
\newblock URL \url{http://arxiv.org/abs/2205.13600}.

\bibitem[Caggiano et~al.(2022{\natexlab{b}})Caggiano, Wang, Durandau, Song,
  Tassa, Sartori, and Kumar]{MyoChallenge2022}
Caggiano, V., Wang, H., Durandau, G., Song, S., Tassa, Y., Sartori, M., and
  Kumar, V.
\newblock Myochallenge: Learning contact-rich manipulation using a
  musculoskeletal hand.
\newblock \url{ https://sites.google.com/view/myochallenge },
  2022{\natexlab{b}}.

\bibitem[Caruana(1997)]{caruana_multitask_nodate}
Caruana, R.
\newblock Multitask learning.
\newblock \emph{Machine learning}, 28:\penalty0 41--75, 1997.

\bibitem[Chen et~al.(2021)Chen, Xu, and Agrawal]{chen_system_2021}
Chen, T., Xu, J., and Agrawal, P.
\newblock A system for general in-hand object re-orientation.
\newblock \emph{{arXiv} preprint {arXiv}:2111.03043}, 2021.

\bibitem[Cheung et~al.(2020)Cheung, Cheung, Zhang, Chan, Ha, Chen, and
  Cheung]{cheung_plasticity_2020}
Cheung, V. C.~K., Cheung, B. M.~F., Zhang, J.~H., Chan, Z. Y.~S., Ha, S. C.~W.,
  Chen, C.-Y., and Cheung, R. T.~H.
\newblock Plasticity of muscle synergies through fractionation and merging
  during development and training of human runners.
\newblock \emph{Nature Communications}, 11\penalty0 (1):\penalty0 4356, 2020.
\newblock ISSN 2041-1723.
\newblock \doi{10.1038/s41467-020-18210-4}.
\newblock URL \url{https://doi.org/10.1038/s41467-020-18210-4}.

\bibitem[Dai et~al.(2016)Dai, He, and Sun]{dai_instance-aware_2016}
Dai, J., He, K., and Sun, J.
\newblock Instance-aware semantic segmentation via multi-task network cascades.
\newblock In \emph{2016 {IEEE} Conference on Computer Vision and Pattern
  Recognition ({CVPR})}, pp.\  3150--3158. {IEEE}, 2016.
\newblock ISBN 978-1-4673-8851-1.
\newblock \doi{10.1109/CVPR.2016.343}.
\newblock URL \url{http://ieeexplore.ieee.org/document/7780712/}.

\bibitem[Dasari et~al.(2023)Dasari, Gupta, and Kumar]{dasari_learning_2023}
Dasari, S., Gupta, A., and Kumar, V.
\newblock Learning dexterous manipulation from exemplar object trajectories and
  pre-grasps.
\newblock In \emph{2023 IEEE International Conference on Robotics and
  Automation (ICRA)}, pp.\  3889--3896. IEEE, 2023.

\bibitem[d'Avella \& Bizzi(2005)d'Avella and Bizzi]{davella_shared_2005}
d'Avella, A. and Bizzi, E.
\newblock Shared and specific muscle synergies in natural motor behaviors.
\newblock \emph{Proceedings of the National Academy of Sciences}, 102\penalty0
  (8):\penalty0 3076--3081, 2005.
\newblock ISSN 0027-8424, 1091-6490.
\newblock \doi{10.1073/pnas.0500199102}.
\newblock URL \url{https://pnas.org/doi/full/10.1073/pnas.0500199102}.

\bibitem[d'Avella et~al.(2003)d'Avella, Saltiel, and
  Bizzi]{davella_combinations_2003}
d'Avella, A., Saltiel, P., and Bizzi, E.
\newblock Combinations of muscle synergies in the construction of a natural
  motor behavior.
\newblock \emph{Nature neuroscience}, 6\penalty0 (3):\penalty0 300--308, 2003.

\bibitem[Delp et~al.(2007)Delp, Anderson, Arnold, Loan, Habib, John,
  Guendelman, and Thelen]{delp_opensim_2007}
Delp, S.~L., Anderson, F.~C., Arnold, A.~S., Loan, P., Habib, A., John, C.~T.,
  Guendelman, E., and Thelen, D.~G.
\newblock {OpenSim}: Open-source software to create and analyze dynamic
  simulations of movement.
\newblock \emph{{IEEE} Transactions on Biomedical Engineering}, 54\penalty0
  (11):\penalty0 1940--1950, 2007.
\newblock \doi{10.1109/TBME.2007.901024}.

\bibitem[Dominici et~al.(2011)Dominici, Ivanenko, Cappellini, d’Avella,
  Mondì, Cicchese, Fabiano, Silei, Paolo, Giannini, Poppele, and
  Lacquaniti]{dominici_locomotor_2011}
Dominici, N., Ivanenko, Y.~P., Cappellini, G., d’Avella, A., Mondì, V.,
  Cicchese, M., Fabiano, A., Silei, T., Paolo, A.~D., Giannini, C., Poppele,
  R.~E., and Lacquaniti, F.
\newblock Locomotor primitives in newborn babies and their development.
\newblock \emph{Science}, 334\penalty0 (6058):\penalty0 997--999, 2011.
\newblock \doi{10.1126/science.1210617}.
\newblock URL \url{https://www.science.org/doi/abs/10.1126/science.1210617}.

\bibitem[Geijtenbeek et~al.(2013)Geijtenbeek, van~de Panne, and van~der
  Stappen]{geijtenbeek_flexible_2013}
Geijtenbeek, T., van~de Panne, M., and van~der Stappen, A.~F.
\newblock Flexible muscle-based locomotion for bipedal creatures.
\newblock \emph{{ACM} Transactions on Graphics}, 32\penalty0 (6):\penalty0
  1--11, 2013.
\newblock ISSN 0730-0301, 1557-7368.
\newblock \doi{10.1145/2508363.2508399}.
\newblock URL \url{https://dl.acm.org/doi/10.1145/2508363.2508399}.

\bibitem[Goyal et~al.(2019)Goyal, Islam, Strouse, Ahmed, Botvinick, Larochelle,
  Bengio, and Levine]{goyal_infobot_2019}
Goyal, A., Islam, R., Strouse, D., Ahmed, Z., Botvinick, M., Larochelle, H.,
  Bengio, Y., and Levine, S.
\newblock {InfoBot}: Transfer and exploration via the information bottleneck,
  2019.
\newblock URL \url{http://arxiv.org/abs/1901.10902}.

\bibitem[Hasenclever et~al.(2020)Hasenclever, Pardo, Hadsell, Heess, and
  Merel]{hasenclever_comic_2020}
Hasenclever, L., Pardo, F., Hadsell, R., Heess, N., and Merel, J.
\newblock {CoMic}: Complementary task learning \& mimicry for reusable skills.
\newblock In \emph{Proceedings of the 37th International Conference on Machine
  Learning}, {ICML}'20. {JMLR}.org, 2020.

\bibitem[Hausman et~al.(2018)Hausman, Springenberg, Wang, Heess, and
  Riedmiller]{hausman_learning_2018}
Hausman, K., Springenberg, J.~T., Wang, Z., Heess, N., and Riedmiller, M.
\newblock Learning an embedding space for transferable robot skills.
\newblock In \emph{International Conference on Learning Representations}, 2018.

\bibitem[Heldstab et~al.(2020)Heldstab, Isler, Schuppli, and van
  Schaik]{heldstab_when_2020}
Heldstab, S.~A., Isler, K., Schuppli, C., and van Schaik, C.~P.
\newblock When ontogeny recapitulates phylogeny: Fixed neurodevelopmental
  sequence of manipulative skills among primates.
\newblock \emph{Science Advances}, 6\penalty0 (30):\penalty0 eabb4685, 2020.
\newblock \doi{10.1126/sciadv.abb4685}.
\newblock URL \url{https://www.science.org/doi/10.1126/sciadv.abb4685}.
\newblock Publisher: American Association for the Advancement of Science.

\bibitem[Ikkala et~al.(2022)Ikkala, Fischer, Klar, Bachinski, Fleig, Howes,
  Hämäläinen, Müller, Murray-Smith, and Oulasvirta]{ikkala_breathing_2022}
Ikkala, A., Fischer, F., Klar, M., Bachinski, M., Fleig, A., Howes, A.,
  Hämäläinen, P., Müller, J., Murray-Smith, R., and Oulasvirta, A.
\newblock Breathing life into biomechanical user models.
\newblock In \emph{Proceedings of the 35th Annual {ACM} Symposium on User
  Interface Software and Technology}, {UIST} '22. Association for Computing
  Machinery, 2022.
\newblock ISBN 978-1-4503-9320-1.
\newblock \doi{10.1145/3526113.3545689}.
\newblock URL \url{https://doi.org/10.1145/3526113.3545689}.
\newblock event-place: Bend, {OR}, {USA}.

\bibitem[Jain et~al.(2019)Jain, Li, Singhal, Rajeswaran, Kumar, and
  Todorov]{jain_learning_2019}
Jain, D., Li, A., Singhal, S., Rajeswaran, A., Kumar, V., and Todorov, E.
\newblock Learning deep visuomotor policies for dexterous hand manipulation.
\newblock In \emph{2019 International Conference on Robotics and Automation
  ({ICRA})}, pp.\  3636--3643, 2019.
\newblock \doi{10.1109/ICRA.2019.8794033}.

\bibitem[Jeannerod(1988)]{jeannerod_neural_1988}
Jeannerod, M.
\newblock \emph{The neural and behavioural organization of goal-directed
  movements}.
\newblock Clarendon Press/Oxford University Press., 1988.

\bibitem[Jiang et~al.(2019)Jiang, Van~Wouwe, De~Groote, and
  Liu]{jiang_synthesis_2019}
Jiang, Y., Van~Wouwe, T., De~Groote, F., and Liu, C.~K.
\newblock Synthesis of biologically realistic human motion using joint torque
  actuation.
\newblock \emph{{ACM} Transactions on Graphics}, 38\penalty0 (4):\penalty0
  1--12, 2019.
\newblock ISSN 0730-0301, 1557-7368.
\newblock \doi{10.1145/3306346.3322966}.
\newblock URL \url{https://dl.acm.org/doi/10.1145/3306346.3322966}.

\bibitem[Joos et~al.(2020)Joos, Péan, and Goksel]{joos_reinforcement_2020}
Joos, E., Péan, F., and Goksel, O.
\newblock Reinforcement learning of musculoskeletal control from functional
  simulations, 2020.
\newblock URL \url{http://arxiv.org/abs/2007.06669}.

\bibitem[Kroemer \& Sukhatme(2016)Kroemer and Sukhatme]{kroemer_learning_2016}
Kroemer, O. and Sukhatme, G.~S.
\newblock Learning relevant features for manipulation skills using meta-level
  priors, 2016.
\newblock URL \url{http://arxiv.org/abs/1605.04439}.

\bibitem[Kumar(2016)]{kumar2016manipulators}
Kumar, V.
\newblock \emph{Manipulators and Manipulation in high dimensional spaces}.
\newblock PhD thesis, University of Washington, 2016.

\bibitem[Kumar et~al.(2016)Kumar, Todorov, and Levine]{kumar_optimal_2016}
Kumar, V., Todorov, E., and Levine, S.
\newblock Optimal control with learned local models: Application to dexterous
  manipulation.
\newblock In \emph{2016 {IEEE} International Conference on Robotics and
  Automation ({ICRA})}, pp.\  378--383, 2016.
\newblock \doi{10.1109/ICRA.2016.7487156}.

\bibitem[Lee et~al.(2015)Lee, Asakawa, Dennerlein, and
  Jindrich]{lee_finger_2015}
Lee, J.~H., Asakawa, D.~S., Dennerlein, J.~T., and Jindrich, D.~L.
\newblock Finger muscle attachments for an {OpenSim} upper-extremity model.
\newblock \emph{{PLOS} {ONE}}, 10\penalty0 (4):\penalty0 e0121712, 2015.
\newblock ISSN 1932-6203.
\newblock \doi{10.1371/journal.pone.0121712}.
\newblock URL \url{https://dx.plos.org/10.1371/journal.pone.0121712}.

\bibitem[Lee et~al.(2018)Lee, Yu, Park, Aanjaneya, Sifakis, and
  Lee]{lee_dexterous_2018}
Lee, S., Yu, R., Park, J., Aanjaneya, M., Sifakis, E., and Lee, J.
\newblock Dexterous manipulation and control with volumetric muscles.
\newblock \emph{{ACM} Transactions on Graphics}, 37\penalty0 (4):\penalty0
  1--13, 2018.
\newblock ISSN 0730-0301, 1557-7368.
\newblock \doi{10.1145/3197517.3201330}.
\newblock URL \url{https://dl.acm.org/doi/10.1145/3197517.3201330}.

\bibitem[Lee et~al.(2019)Lee, Park, Lee, and Lee]{lee_scalable_2019}
Lee, S., Park, M., Lee, K., and Lee, J.
\newblock Scalable muscle-actuated human simulation and control.
\newblock \emph{{ACM} Transactions on Graphics}, 38\penalty0 (4):\penalty0
  1--13, 2019.
\newblock ISSN 0730-0301, 1557-7368.
\newblock \doi{10.1145/3306346.3322972}.
\newblock URL \url{https://dl.acm.org/doi/10.1145/3306346.3322972}.

\bibitem[Liu \& Hodgins(2018)Liu and Hodgins]{liu_learning_2018}
Liu, L. and Hodgins, J.
\newblock Learning basketball dribbling skills using trajectory optimization
  and deep reinforcement learning.
\newblock \emph{{ACM} Transactions on Graphics}, 37\penalty0 (4):\penalty0
  1--14, 2018.
\newblock ISSN 0730-0301, 1557-7368.
\newblock \doi{10.1145/3197517.3201315}.
\newblock URL \url{https://dl.acm.org/doi/10.1145/3197517.3201315}.

\bibitem[Liu et~al.(2019)Liu, Johns, and Davison]{liu_end--end_2019}
Liu, S., Johns, E., and Davison, A.~J.
\newblock End-to-end multi-task learning with attention.
\newblock In \emph{2019 {IEEE}/{CVF} Conference on Computer Vision and Pattern
  Recognition ({CVPR})}, pp.\  1871--1880. {IEEE}, 2019.
\newblock ISBN 978-1-72813-293-8.
\newblock \doi{10.1109/CVPR.2019.00197}.
\newblock URL \url{https://ieeexplore.ieee.org/document/8954221/}.

\bibitem[{McFarland} et~al.(2021){McFarland}, Binder-Markey, Nichols, Wohlman,
  de~Bruin, and Murray]{mcfarland_musculoskeletal_2021}
{McFarland}, D.~C., Binder-Markey, B.~I., Nichols, J.~A., Wohlman, S.~J.,
  de~Bruin, M., and Murray, W.~M.
\newblock A musculoskeletal model of the hand and wrist capable of simulating
  functional tasks.
\newblock \emph{{bioRxiv}}, pp.\  2021.12.28.474357, 2021.
\newblock \doi{10.1101/2021.12.28.474357}.
\newblock URL
  \url{http://biorxiv.org/content/early/2021/12/30/2021.12.28.474357.abstract}.

\bibitem[Merel et~al.(2017)Merel, Tassa, {TB}, Srinivasan, Lemmon, Wang, Wayne,
  and Heess]{merel_learning_2017}
Merel, J., Tassa, Y., {TB}, D., Srinivasan, S., Lemmon, J., Wang, Z., Wayne,
  G., and Heess, N.
\newblock Learning human behaviors from motion capture by adversarial
  imitation, 2017.
\newblock URL \url{http://arxiv.org/abs/1707.02201}.

\bibitem[Merel et~al.(2019)Merel, Hasenclever, Galashov, Ahuja, Pham, Wayne,
  Teh, and Heess]{merel_neural_2019}
Merel, J., Hasenclever, L., Galashov, A., Ahuja, A., Pham, V., Wayne, G., Teh,
  Y.~W., and Heess, N.
\newblock Neural probabilistic motor primitives for humanoid control, 2019.
\newblock URL \url{http://arxiv.org/abs/1811.11711}.

\bibitem[Nagabandi et~al.(2019)Nagabandi, Konoglie, Levine, and
  Kumar]{nagabandi_deep_2019}
Nagabandi, A., Konoglie, K., Levine, S., and Kumar, V.
\newblock Deep dynamics models for learning dexterous manipulation, 2019.
\newblock URL \url{http://arxiv.org/abs/1909.11652}.

\bibitem[Park et~al.(2022)Park, Min, Chang, Lee, Park, and
  Lee]{park_generative_2022}
Park, J., Min, S., Chang, P.~S., Lee, J., Park, M., and Lee, J.
\newblock Generative {GaitNet}, 2022.
\newblock URL \url{http://arxiv.org/abs/2201.12044}.

\bibitem[Pastor et~al.(2012)Pastor, Kalakrishnan, Righetti, and
  Schaal]{pastor_towards_2012}
Pastor, P., Kalakrishnan, M., Righetti, L., and Schaal, S.
\newblock Towards associative skill memories.
\newblock \emph{2012 12th {IEEE}-{RAS} International Conference on Humanoid
  Robots (Humanoids 2012)}, pp.\  309--315, 2012.
\newblock \doi{10.1109/HUMANOIDS.2012.6651537}.
\newblock URL \url{http://ieeexplore.ieee.org/document/6651537/}.
\newblock Conference Name: 2012 12th {IEEE}-{RAS} International Conference on
  Humanoid Robots (Humanoids 2012) {ISBN}: 9781467313698 Place: Osaka, Japan
  Publisher: {IEEE}.

\bibitem[Raffin et~al.(2021)Raffin, Hill, Gleave, Kanervisto, Ernestus, and
  Dormann]{raffin_stable-baselines3_2021}
Raffin, A., Hill, A., Gleave, A., Kanervisto, A., Ernestus, M., and Dormann, N.
\newblock Stable-baselines3: Reliable reinforcement learning implementations.
\newblock \emph{Journal of Machine Learning Research}, 22\penalty0
  (268):\penalty0 1--8, 2021.
\newblock URL \url{http://jmlr.org/papers/v22/20-1364.html}.

\bibitem[Rajeswaran et~al.(2018)Rajeswaran, Kumar, Gupta, Vezzani, Schulman,
  Todorov, and Levine]{rajeswaran_learning_2018}
Rajeswaran, A., Kumar, V., Gupta, A., Vezzani, G., Schulman, J., Todorov, E.,
  and Levine, S.
\newblock Learning complex dexterous manipulation with deep reinforcement
  learning and demonstrations.
\newblock In \emph{Proceedings of Robotics: Science and Systems ({RSS})}, 2018.

\bibitem[Rong et~al.(2021)Rong, Shiratori, and Joo]{rong2021frankmocap}
Rong, Y., Shiratori, T., and Joo, H.
\newblock Frankmocap: A monocular 3d whole-body pose estimation system via
  regression and integration.
\newblock In \emph{Proceedings of the IEEE/CVF International Conference on
  Computer Vision}, pp.\  1749--1759, 2021.

\bibitem[R{\"u}ckert \& d'Avella(2013)R{\"u}ckert and
  d'Avella]{ruckert_learned_2013}
R{\"u}ckert, E. and d'Avella, A.
\newblock Learned parametrized dynamic movement primitives with shared
  synergies for controlling robotic and musculoskeletal systems.
\newblock \emph{Frontiers in computational neuroscience}, 7:\penalty0 138,
  2013.

\bibitem[Rueckert et~al.(2015)Rueckert, Mundo, Paraschos, Peters, and
  Neumann]{rueckert_extracting_2015}
Rueckert, E., Mundo, J., Paraschos, A., Peters, J., and Neumann, G.
\newblock Extracting low-dimensional control variables for movement primitives.
\newblock In \emph{2015 IEEE International Conference on Robotics and
  Automation (ICRA)}, pp.\  1511--1518. IEEE, 2015.

\bibitem[Santello et~al.(2002)Santello, Flanders, and
  Soechting]{santello_patterns_2002}
Santello, M., Flanders, M., and Soechting, J.~F.
\newblock Patterns of hand motion during grasping and the influence of sensory
  guidance.
\newblock \emph{Journal of Neuroscience}, 22\penalty0 (4):\penalty0 1426--1435,
  2002.

\bibitem[Saul et~al.(2015)Saul, Hu, Goehler, Vidt, Daly, Velisar, and
  Murray]{saul_benchmarking_2015}
Saul, K.~R., Hu, X., Goehler, C.~M., Vidt, M.~E., Daly, M., Velisar, A., and
  Murray, W.~M.
\newblock Benchmarking of dynamic simulation predictions in two software
  platforms using an upper limb musculoskeletal model.
\newblock \emph{Computer Methods in Biomechanics and Biomedical Engineering},
  18\penalty0 (13):\penalty0 1445--1458, 2015.
\newblock \doi{10.1080/10255842.2014.916698}.
\newblock URL \url{https://doi.org/10.1080/10255842.2014.916698}.

\bibitem[Schulman et~al.(2017)Schulman, Wolski, Dhariwal, Radford, and
  Klimov]{schulman_proximal_2017}
Schulman, J., Wolski, F., Dhariwal, P., Radford, A., and Klimov, O.
\newblock Proximal policy optimization algorithms, 2017.
\newblock URL \url{http://arxiv.org/abs/1707.06347}.

\bibitem[Schumacher et~al.(2022)Schumacher, Häufle, Büchler, Schmitt, and
  Martius]{schumacher_dep-rl_2022}
Schumacher, P., Häufle, D., Büchler, D., Schmitt, S., and Martius, G.
\newblock {DEP}-{RL}: Embodied exploration for reinforcement learning in
  overactuated and musculoskeletal systems, 2022.
\newblock URL \url{https://arxiv.org/abs/2206.00484}.

\bibitem[Seth et~al.(2018)Seth, Hicks, Uchida, Habib, Dembia, Dunne, Ong,
  {DeMers}, Rajagopal, Millard, Hamner, Arnold, Yong, Lakshmikanth, Sherman,
  Ku, and Delp]{seth_opensim_2018}
Seth, A., Hicks, J.~L., Uchida, T.~K., Habib, A., Dembia, C.~L., Dunne, J.~J.,
  Ong, C.~F., {DeMers}, M.~S., Rajagopal, A., Millard, M., Hamner, S.~R.,
  Arnold, E.~M., Yong, J.~R., Lakshmikanth, S.~K., Sherman, M.~A., Ku, J.~P.,
  and Delp, S.~L.
\newblock {OpenSim}: Simulating musculoskeletal dynamics and neuromuscular
  control to study human and animal movement.
\newblock \emph{{PLOS} Computational Biology}, 14:\penalty0 1--20, 2018.
\newblock \doi{10.1371/journal.pcbi.1006223}.
\newblock URL \url{https://doi.org/10.1371/journal.pcbi.1006223}.

\bibitem[Sobinov \& Bensmaia(2021)Sobinov and Bensmaia]{sobinov_neural_2021}
Sobinov, A.~R. and Bensmaia, S.~J.
\newblock The neural mechanisms of manual dexterity.
\newblock \emph{Nature Reviews Neuroscience}, 22\penalty0 (12):\penalty0
  741--757, 2021.
\newblock ISSN 1471-0048.
\newblock \doi{10.1038/s41583-021-00528-7}.
\newblock URL \url{https://doi.org/10.1038/s41583-021-00528-7}.

\bibitem[Song et~al.(2020)Song, Kidziński, Peng, Ong, Hicks, Levine, Atkeson,
  and Delp]{song_deep_2020}
Song, S., Kidziński, {\textbackslash}., Peng, X.~B., Ong, C., Hicks, J.,
  Levine, S., Atkeson, C.~G., and Delp, S.~L.
\newblock Deep reinforcement learning for modeling human locomotion control in
  neuromechanical simulation.
\newblock \emph{{bioRxiv}}, 2020.
\newblock \doi{10.1101/2020.08.11.246801}.

\bibitem[Sun et~al.(2020)Sun, Panda, Feris, and Saenko]{sun_adashare_2020}
Sun, X., Panda, R., Feris, R., and Saenko, K.
\newblock {AdaShare}: learning what to share for efficient deep multi-task
  learning.
\newblock In \emph{Proceedings of the 34th International Conference on Neural
  Information Processing Systems}, {NIPS}'20, pp.\  8728--8740. Curran
  Associates Inc., 2020.
\newblock ISBN 978-1-71382-954-6.

\bibitem[Sutton \& Barto(2018)Sutton and Barto]{sutton_reinforcement_2018}
Sutton, R.~S. and Barto, A.~G.
\newblock \emph{Reinforcement Learning: An Introduction}.
\newblock The {MIT} Press, second edition, 2018.
\newblock URL \url{http://incompleteideas.net/book/the-book-2nd.html}.

\bibitem[Taheri et~al.(2020)Taheri, Ghorbani, Black, and
  Tzionas]{taheri_grab_2020-1}
Taheri, O., Ghorbani, N., Black, M.~J., and Tzionas, D.
\newblock {GRAB}: A dataset of whole-body human grasping of objects.
\newblock In Vedaldi, A., Bischof, H., Brox, T., and Frahm, J.-M. (eds.),
  \emph{Computer Vision – {ECCV} 2020}, pp.\  581--600. Springer
  International Publishing, 2020.
\newblock ISBN 978-3-030-58548-8.

\bibitem[Todorov et~al.(2012)Todorov, Erez, and Tassa]{todorov_mujoco_2012}
Todorov, E., Erez, T., and Tassa, Y.
\newblock Mujoco: A physics engine for model-based control.
\newblock In \emph{2012 {IEEE}/{RSJ} International Conference on Intelligent
  Robots and Systems}, pp.\  5026--5033. {IEEE}, 2012.

\bibitem[Tresch et~al.(2006)Tresch, Cheung, and d'Avella]{tresch_matrix_2006}
Tresch, M.~C., Cheung, V. C.~K., and d'Avella, A.
\newblock Matrix factorization algorithms for the identification of muscle
  synergies: Evaluation on simulated and experimental data sets.
\newblock \emph{Journal of Neurophysiology}, 95\penalty0 (4):\penalty0
  2199--2212, 2006.
\newblock ISSN 0022-3077, 1522-1598.
\newblock \doi{10.1152/jn.00222.2005}.
\newblock URL \url{https://www.physiology.org/doi/10.1152/jn.00222.2005}.

\bibitem[Vaswani et~al.(2017)Vaswani, Shazeer, Parmar, Uszkoreit, Jones, Gomez,
  Kaiser, and Polosukhin]{vaswani_attention_2017}
Vaswani, A., Shazeer, N., Parmar, N., Uszkoreit, J., Jones, L., Gomez, A.~N.,
  Kaiser, L., and Polosukhin, I.
\newblock Attention is all you need.
\newblock In Guyon, I., Luxburg, U.~V., Bengio, S., Wallach, H., Fergus, R.,
  Vishwanathan, S., and Garnett, R. (eds.), \emph{Advances in Neural
  Information Processing Systems}, volume~30. Curran Associates, Inc., 2017.
\newblock URL
  \url{https://proceedings.neurips.cc/paper/2017/file/3f5ee243547dee91fbd053c1c4a845aa-Paper.pdf}.

\bibitem[Wang et~al.(2022)Wang, Caggiano, Durandau, Sartori, and
  {Vikash}]{wang_myosim_2022}
Wang, H., Caggiano, V., Durandau, G., Sartori, Massimo, K., and {Vikash}.
\newblock {MyoSim}: Fast and physiologically realistic {MuJoCo} models for
  musculoskeletal and exoskeletal studies.
\newblock In \emph{2022 {IEEE} international conference on robotics and
  automation ({ICRA})}. {IEEE}, 2022.

\bibitem[Wang et~al.(2012)Wang, Hamner, Delp, and Koltun]{wang_optimizing_2012}
Wang, J.~M., Hamner, S.~R., Delp, S.~L., and Koltun, V.
\newblock Optimizing locomotion controllers using biologically-based actuators
  and objectives.
\newblock \emph{{ACM} Transactions on Graphics}, 31\penalty0 (4):\penalty0
  1--11, 2012.
\newblock ISSN 0730-0301, 1557-7368.
\newblock \doi{10.1145/2185520.2185521}.
\newblock URL \url{https://dl.acm.org/doi/10.1145/2185520.2185521}.

\bibitem[Won et~al.(2021)Won, Gopinath, and Hodgins]{won_control_2021}
Won, J., Gopinath, D., and Hodgins, J.
\newblock Control strategies for physically simulated characters performing
  two-player competitive sports.
\newblock \emph{{ACM} Transactions on Graphics}, 40\penalty0 (4):\penalty0
  1--11, 2021.
\newblock ISSN 0730-0301, 1557-7368.
\newblock \doi{10.1145/3450626.3459761}.
\newblock URL \url{https://dl.acm.org/doi/10.1145/3450626.3459761}.

\bibitem[Yang et~al.(2019)Yang, Logan, and Giszter]{yang_motor_2019}
Yang, Q., Logan, D., and Giszter, S.~F.
\newblock Motor primitives are determined in early development and are then
  robustly conserved into adulthood.
\newblock \emph{Proceedings of the National Academy of Sciences}, 116\penalty0
  (24):\penalty0 12025--12034, 2019.
\newblock \doi{10.1073/pnas.1821455116}.
\newblock URL \url{https://www.pnas.org/doi/abs/10.1073/pnas.1821455116}.

\bibitem[Yin et~al.(2021)Yin, Yang, Van De~Panne, and
  Yin]{yin_discovering_2021}
Yin, Z., Yang, Z., Van De~Panne, M., and Yin, K.
\newblock Discovering diverse athletic jumping strategies.
\newblock \emph{{ACM} Transactions on Graphics}, 40\penalty0 (4):\penalty0
  1--17, 2021.
\newblock ISSN 0730-0301, 1557-7368.
\newblock \doi{10.1145/3450626.3459817}.
\newblock URL \url{https://dl.acm.org/doi/10.1145/3450626.3459817}.

\bibitem[Zhang \& Yeung(2014)Zhang and Yeung]{zhang_regularization_2014}
Zhang, Y. and Yeung, D.-Y.
\newblock A regularization approach to learning task relationships in multitask
  learning.
\newblock \emph{{ACM} Transactions on Knowledge Discovery from Data},
  8\penalty0 (3):\penalty0 12:1--12:31, 2014.
\newblock ISSN 1556-4681.
\newblock \doi{10.1145/2538028}.
\newblock URL \url{https://doi.org/10.1145/2538028}.

\end{thebibliography}
\bibliographystyle{icml2023}

\newpage
\appendix
\counterwithin{figure}{section}
\counterwithin{table}{section}

\onecolumn
\newpage

\section{Appendix}

\begin{figure}[h]
     \centering
         \centering
         \includegraphics[width=0.45\textwidth]{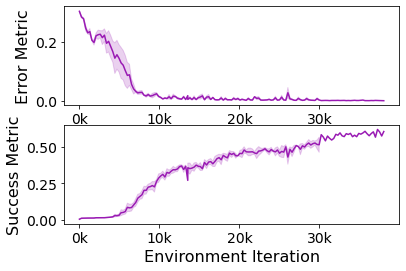}
         \caption{Success and error metrics for the multi-task policy trained on 14 environments in the first 40k iterations on 4 seeds (line average and errors as shaded area).}
          \label{fig:Multi-Task_training_curve_zoom_in}
\end{figure}

\begin{figure}[h!]
 \centering
 \includegraphics[width=.5\textwidth]{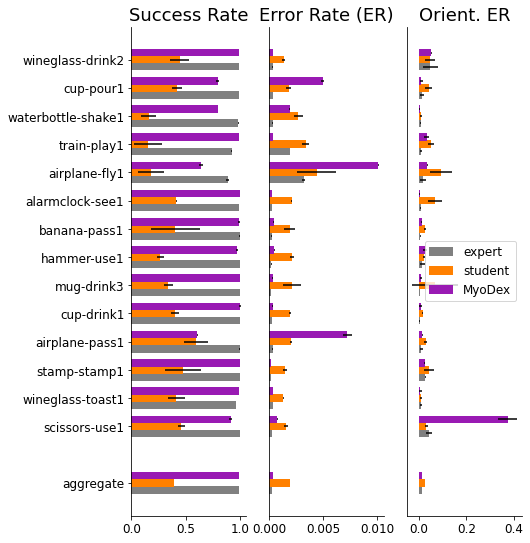}
     \caption{\textbf{Baselines}: Success, error rate, orientation error metrics (in the left, middle and right columns, respectively -- see definitions in Sec. \ref{sec:metrics}) for Individual-Task Experts $\pi_i$, Multi-task Student policy $\pi^*$, and Multi-task \name~ $\pi^\#$ policy. On the Y-axis the 14 tasks used to traing \name{} are reported, in addition to an aggregate information. \name{} is able to match individual-Task Experts solutions across the 3 differnt metrics. Nevertheless, the multi-task student policy was able to achieve lower perforances overall in most of the individual tasks.
     }
    
     \label{fig:MetricsDistilled_in-domain}
\end{figure}

\begin{figure}[h!]
     \centering
     \includegraphics[width=.4\textwidth]{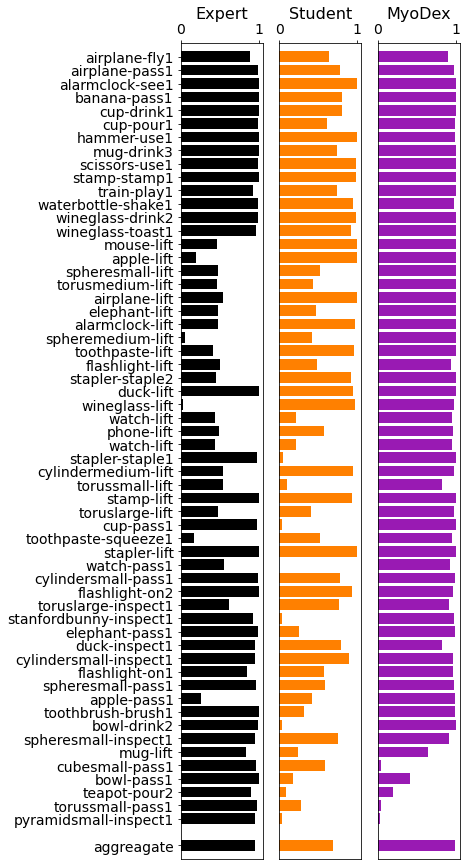}
     \caption{\textbf{Summary of all Tasks.} Left column tasks solved by single expert policies. Right columns, task fine tuning based on \name{}. See also Table \ref{table:FineTuning}. Training values reported at $12.5k$ iterations.}
     \label{fig:alltasks}
     
\end{figure}

\begin{table}[h]
    \begin{center}
        \begin{tabular}{ |c|c| } 
         \hline
         Samples for Iterations & $4096$ \\ 
         Discount Factor ($\gamma$) & $0.95$ \\ 
         GAE-$\lambda$ & $0.95$ \\
         VF Coefficient (c1) & $0.5$ \\ 
         Entropy Bonus (c2) & $0.001$ \\ 
         Clip Parameter ($\epsilon$)  &  $0.2$\\ 
         Batch Size & $256$\\ 
         Epochs & $5$\\ 
         Network Size & $ pi =[256, 128], vf = [256, 128] $ \\
         \hline
        \end{tabular}
        \caption{Parameters adopted for the reinforcement learning models.}
        \label{Table:Parameters}
    \end{center}
% \vspace{3cm}
\end{table}

\subsection{Imitation learning.}\label{section:imitation_learning}
In addition to \name{} $\pi^\#$, we also train a baseline agent using $\pi^*$ expert-student method \cite{jain_learning_2019, chen_system_2021}. Individual task-specific policies ($\pi_i$) were used as experts. We developed a dataset with 1M samples of observation-action tuples for each of those policies. A neural network was trained via supervised learning to learn the association between observations and actions to obtain a single policy $\pi^*(a_t| s_t$) capable of multiple task behaviors.

For distilling the single expert agents into one, a neural network of the same size of the single agent was used. We adopted a batch size of 256, and Adadelta optimizer with a learning rate of $0.25$, a Discount Factor ($\gamma$) of $0.995$, and 10 epochs.

\begin{figure}[h]
     \centering
     \includegraphics[width=.4\textwidth]{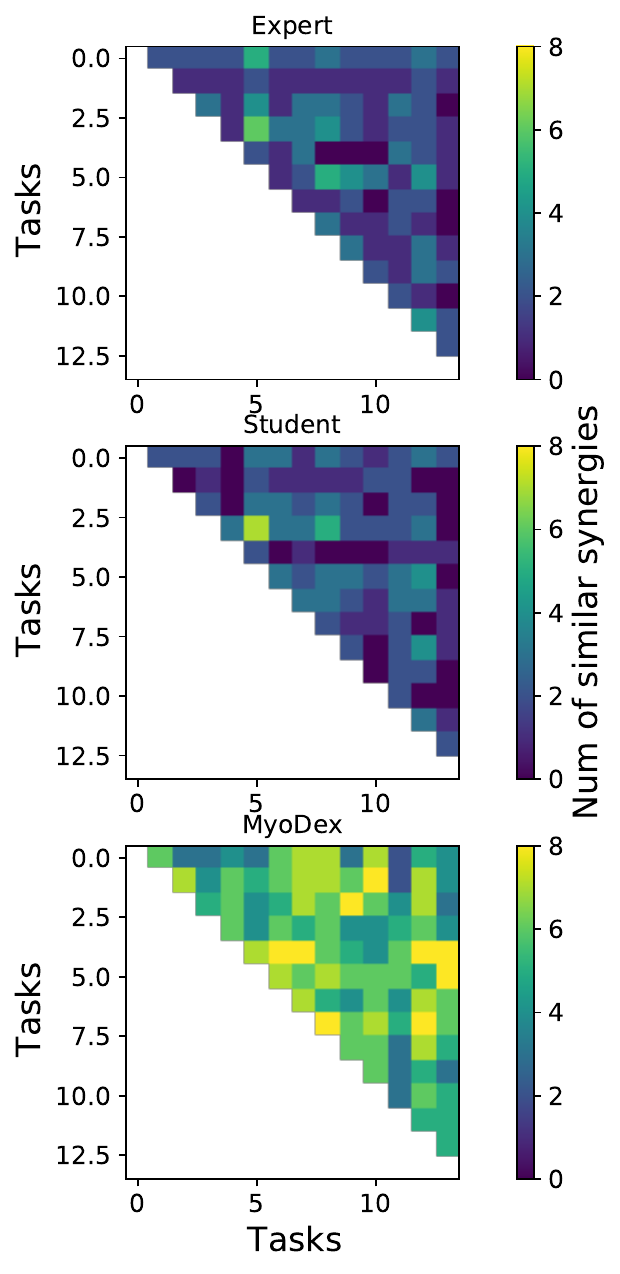}
     \caption{Cosine Similarity between 12 synergies extracted from 14 different tasks. Top - expert policies. Middle - student policy. Bottom -- \name{}. On average the number of similar synergies for expert, student, \name~ (mean +/- std over 10 repetitions with different random seeds) was $1.86\pm1.19$, $1.71\pm1.29$ and $5.56\pm1.59$, respectively.}
     \label{Fig:CosSimilarity}
\end{figure}

\begin{table*}[h]
\small
 % \vspace{-1.5cm}
    \centering
        \begin{tabular}{|l|c|c|c|c|c|}
        \hline
        \textbf{Task} &  \multicolumn{3}{|c|}{\textbf{Multi-task Success}}   &  \multicolumn{2}{|c|}{\textbf{ Iter. to reach Success of 0.8}} \\
  & \textbf{ @ 1k  Iter.} &   @ \textbf{2k  Iter.} &   @ \textbf{3k  Iter.} &   \textbf{Multi-Task} &   \textbf{Expert} \\
        \hline
stamp-stamp1           &      0.981538 &      0.997949 &      1.000000 &               247  &                 3458  \\
banana-pass1           &      0.910000 &      0.993333 &      0.998571 &               247  &                 4446  \\
cup-drink1             &      0.991724 &      0.998161 &      0.977471 &               247  &                 3952  \\
mug-drink3             &      0.978667 &      0.999467 &      1.000000 &               247  &                 3458  \\
alarmclock-see1        &      0.984444 &      0.997778 &      1.000000 &               494  &                 4940  \\
train-play1            &      0.822278 &      0.929114 &      0.987848 &               741  &                 8398  \\
scissors-use1          &      0.754699 &      0.945542 &      0.986988 &              1235  &                 5434  \\
wineglass-drink2       &      0.714943 &      0.924138 &      0.985287 &              1235  &                 4446  \\
hammer-use1            &      0.781429 &      0.870000 &      0.972857 &              1482  &                 3952  \\
wineglass-toast1       &      0.713846 &      0.796410 &      0.902051 &              2223  &                 4199  \\
cup-pour1              &      0.743429 &      0.730857 &      0.830286 &              2964  &                 4446  \\
waterbottle-shake1     &      0.574595 &      0.709189 &      0.743784 &              3458  &                 5434  \\
airplane-fly1          &      0.564675 &      0.606753 &      0.631169 &             12350  &                 7657  \\
airplane-pass1         &      0.436322 &      0.497011 &      0.509425 &             12597  &                 6669  \\
\hline
mouse-lift             &      1.000000 &      1.000000 &      1.000000 &               247  &                     -   \\
apple-lift             &      1.000000 &      1.000000 &      1.000000 &               247  &                     -   \\
spheresmall-lift       &      0.986667 &      1.000000 &      1.000000 &               247  &                     -   \\
torusmedium-lift       &      0.980571 &      1.000000 &      1.000000 &               247  &                     -   \\
airplane-lift          &      0.995122 &      1.000000 &      1.000000 &               247  &                     -   \\
elephant-lift          &      1.000000 &      1.000000 &      1.000000 &               247  &                     -   \\
alarmclock-lift        &      1.000000 &      1.000000 &      1.000000 &               247  &                     -   \\
spheremedium-lift      &      0.998947 &      1.000000 &      1.000000 &               494  &                     -   \\
toothpaste-lift        &      0.971818 &      0.952727 &      0.990000 &               494  &                     -   \\
flashlight-lift        &      0.941714 &      0.942857 &      0.942857 &               494  &                     -   \\
stapler-staple2        &      0.991529 &      1.000000 &      0.999529 &               494  &                     -   \\
duck-lift              &      0.994737 &      1.000000 &      1.000000 &               494  &                     5434   \\
wineglass-lift         &      0.933000 &      0.979500 &      0.980000 &               494  &                     -   \\
watch-lift             &      0.925333 &      0.955556 &      0.955556 &               741  &                     -   \\
phone-lift             &      0.960000 &      0.967742 &      0.967742 &               741  &                     -   \\
stapler-staple1        &      0.893809 &      0.989524 &      0.996190 &               988  &                 5187  \\
cylindermedium-lift    &      0.841111 &      0.970000 &      0.972222 &               988  &                     -   \\
torussmall-lift        &      0.690285  &    0.931428  &      0.915428               &               1235  &      -   \\
stamp-lift             &      0.709756 &      0.980488 &      0.992195 &              1235  &                 3211  \\
toruslarge-lift        &      0.707273 &      0.965455 &      0.977273 &              1235  &                     -   \\
cup-pass1              &      0.609048 &      0.995238 &      1.000000 &              1235  &                 4446  \\
toothpaste-squeeze1    &      0.598421 &      0.943157 &      0.977368 &             1482  &                     -   \\
stapler-lift           &      0.650732 &      0.868293 &      0.982439 &              1482  &                     2717   \\
watch-pass1            &      0.492593 &      0.887407 &      0.884444 &              1729  &                     -   \\
cylindersmall-pass1    &      0.571200 &      0.826667 &      0.901333 &              1976  &                 4693  \\

flashlight-on2         &      0.168791 &      0.695385 &      0.920000 &              2470 &                 6175 \\
toruslarge-inspect1    &      0.251852 &      0.645926 &      0.817778 &              2470  &                     -   \\
stanfordbunny-inspect1 &      0.289157 &      0.591325 &      0.921446 &              2470  &                 6422  \\
elephant-pass1         &      0.506667 &      0.621235 &      0.834568 &              2964 &                 5928 \\
duck-inspect1          &      0.621299 &      0.624935 &      0.820260 &              2964 &                 5681 \\
cylindersmall-inspect1 &      0.420000 &      0.713333 &      0.639444 &              3705  &                 6422  \\
flashlight-on1         &      0.234483 &      0.541609 &      0.626207 &              4446  &                10127  \\
spheresmall-pass1      &      0.191905 &      0.351905 &      0.674286 &              4446  &                 5928  \\
apple-pass1            &      0.344198 &      0.481481 &      0.583210 &              5187  &                     -   \\
toothbrush-brush1      &      0.119375 &      0.353125 &      0.589063 &              5434  &                 4199  \\
bowl-drink2            &      0.075714 &      0.089524 &      0.163810 &              7657  &                 4693  \\
spheresmall-inspect1   &      0.235676 &      0.332432 &      0.438919 &              8151 &                 7163 \\
mug-lift               &      0.326575 &      0.335342 &      0.397808 &                  -   &                 7904  \\
cubesmall-pass1        &      0.024691 &      0.024691 &      0.024691 &                  -   &                 5928  \\
bowl-pass1             &      0.114430 &      0.153418 &      0.184810 &                  -   &                 7163  \\
teapot-pour2           &      0.137627 &      0.150508 &      0.162712 &                  -   &                 7657  \\
torussmall-pass1       &      0.038987 &      0.037975 &      0.037975 &                  -   &                 5928  \\
pyramidsmall-inspect1  &      0.028571 &      0.033333 &      0.035238 &                  -   &                 5187  \\
\hline
\end{tabular}
 \caption{\textbf{\name{} based fine-tuning and expert solutions for all 57 tasks.} Expert solutions could reliably reach 0.80 success for the first 14 tasks but in many other cases they were not able to. A few exceptions at the bottom show success only for expert solutions. We indicated with '-' the lack of success in achieving the success threshold. The first 3 columns report the success rate respectively at $1k$, $2k$ and $3k$ iterations. The 4th and 5th columns, document the iterations at 0.80 success for \name{} based fine-tuning and experts.}
    \label{table:FineTuning}
\end{table*}

\begin{figure}[h]
     \centering
     \includegraphics[width=.4\textwidth]{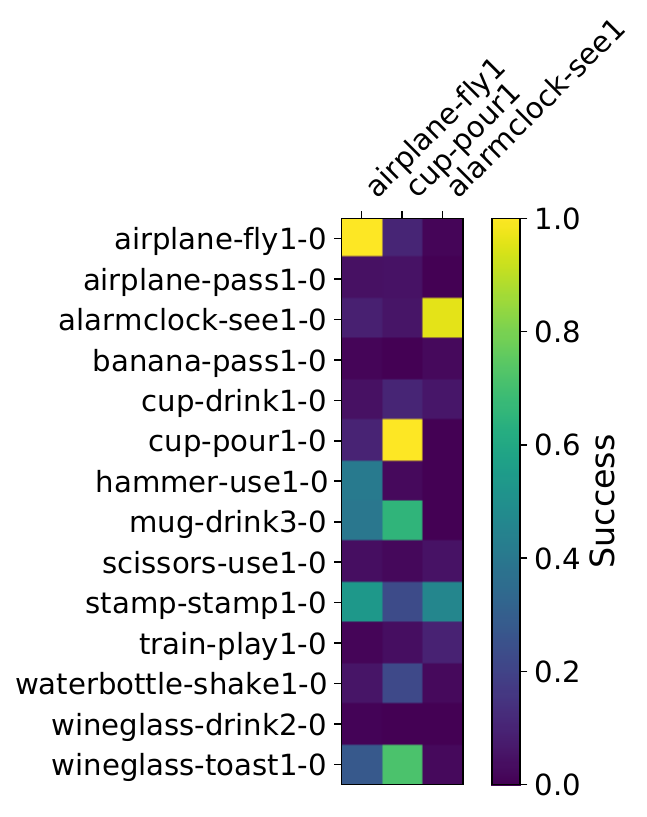}
     \caption{\textbf{Fine-tuning based on expert policies.} Success rate fine-tuning experts solutions (columns) on 14 different environments. This matrix shows that the combination of pre-grasps and the initialization on a pre-trained task is not enough to generalize to new tasks.}
     \label{Fig:FineTuning_Experts}
     % \vspace{3cm}
\end{figure}

\subsection{Noise}\label{section:Noise}

\begin{figure}
     \centering     
     \includegraphics[width=0.4\textwidth]{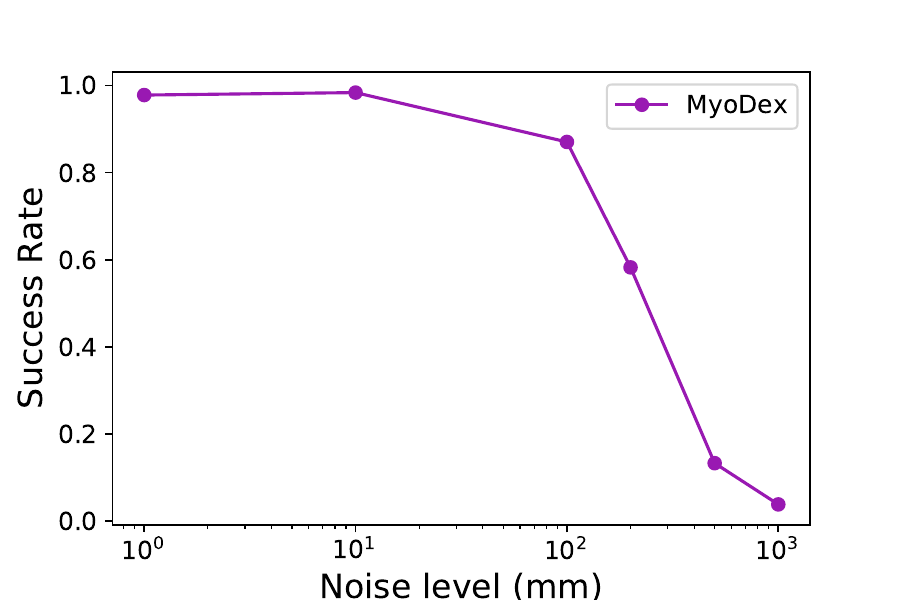}
     \caption{Performance according to addional noise (in mm) in the observation of the object. }
     \label{fig:Noise}
\end{figure}

In real-world scenarios, calculating the precise position, and trajectory of an object is often subject to errors. To address this issue, we conducted a study to investigate the resilience of these policies to noisy measurements.  We emulate real-world tracking errors by gradually adding increasing levels of noise to the trained policies during deployment (see Figure \ref{fig:Noise}). We see that \name{} policies are able to handle significant levels of noise (up to 100mm) in the observation of the object's position with limited loss in performance.

\subsection{\name{} Alternatives}\label{section:MyoDexAlternatives}
\begin{figure}
    \centering
    \includegraphics[width=0.5\textwidth]{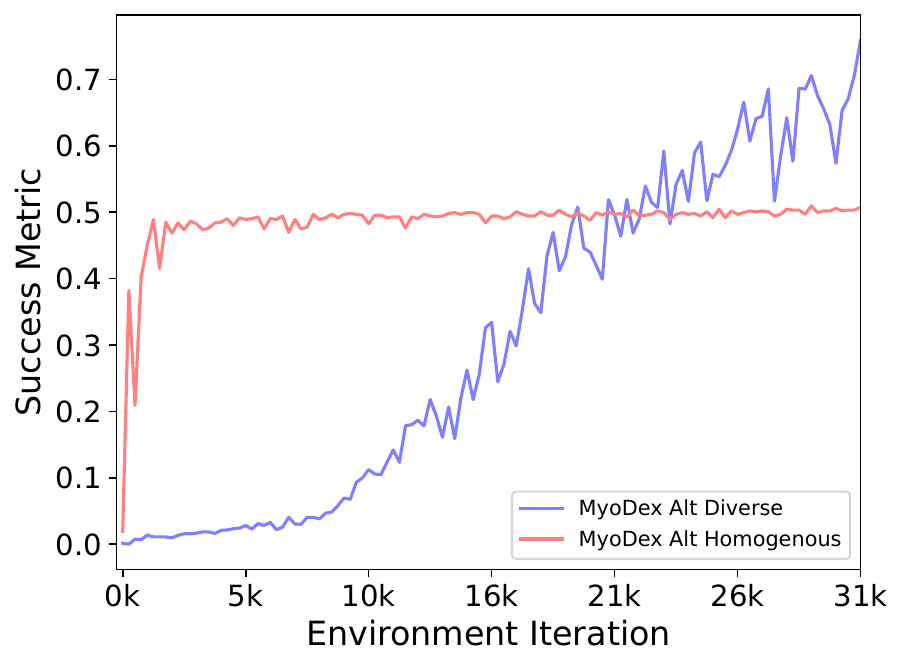}
    \caption{ \textbf{Effect of pre-training task distribution on \name{} training.} 
    \textit{MyoDex Alt Homogenous} shows a quick rise in performance which saturates due to overfitting. In contrast, \textit{MyoDex Alt Diverse} observes a slow start but is gradually able to improve its performance over time. }
    \label{fig:TaskDistr_PreTraining}
\end{figure}

The choice of objects and tasks can significantly impact the effectiveness of the \name{} representation. To investigate this, we conducted an ablation experiment using two new sets of 14 tasks each: a diverse tasks collection  with similar complexity as \name{} --  \textit{MyoDex Alt Diverse} (shown in blue in Figure \ref{fig:Task_Complexity})-- and the other with a homogenous task collection \textit{MyoDex Alt Homogenous} (shown in red in Figure \ref{fig:Task_Complexity}). 
% We then fine-tuned the system on a complementary set of 43 tasks. 
\textit{MyoDex Alt Homogenous} shows a quick rise in performance which saturates due to overfitting \ref{fig:TaskDistr_PreTraining}. In contrast, \textit{MyoDex Alt Diverse} observes a slow start but is gradually able to improve its performance over time. Figure \ref{fig:Success_MyoDexAlternatives} shows that the priors implicitly induced by the richer task diversity leads to better generalization and transfer to new unseen tasks.

\subsection{Synergy probing}
% % Outline how we calculate and eval for synergies
\label{section:muscle_synergies}

To quantify the level of muscle coordination required for accomplishing a given task, we calculated muscle synergies by means of Non-Negative Matrix factorization (NNMF) \cite{tresch_matrix_2006}. 

After training, we played policies for 5 roll-outs to solve specific tasks and we stored the muscle activations (value between 0 and 1) required. Then, a matrix $A$ of muscle activations over time (dimension 39 muscle x total task duration) was fed into a non-negative matrix decomposition (\textit{sklearn}) method.
 
The NNMF method finds two matrices $W$ and $H$ that are respectively the coefficients and the basis vectors which product approximates $A$. Muscle synergies identified by NNMF capture the spatial regularities on the muscle activations whose linear combination minimize muscle reconstruction \cite{bizzi_neural_2013}. This method reveals the amount of variance explained by each of the components. We calculated the Variance Accounted For (VAF) as:
\begin{equation}
    VAF = 100 \cdot \left ( 1 - \frac{(A - W \cdot  H)^2}{A^2} \right ) 
    \label{Eq:VAF}
\end{equation}

Similarity of synergies between two different tasks was calculated using cosine similarity (CS) such as:
$ CS = w_i  \cdot w_j $, where $[w_i, \ w_j] \in W$ are synergy coefficients respectively for the task $i$ and $j$. We used then a threshold of $0.8$ to indicate that 2 synergies were similar Appendix-\ref{Fig:CosSimilarity}.

While the student policy -- obtained with imitation learning -- produced muscle activations similar 
to that of the respective task expert 
but it effectiveness was quite low in task metrics. coordination?

\begin{figure}[h!]
    \centering
    \includegraphics[width=.5\textwidth]{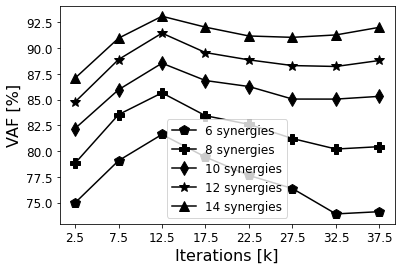}
    \caption{\textbf{Muscle Synergies over learning iterations for the joint multi-task policy.} Variance of the muscle activations (see Sec. \ref{section:muscle_synergies}) explained as function of the number of synergies at different steps of the learning process. }
    \label{fig:Synergies_iterations}
\end{figure}

\subsubsection{Does \name~ produce reusable synergies?} \label{sec:result_muscle_coord}

Biological systems simplify the problem to control the redundant and complex muscuolokeletal systems by resorting on activating particular muscle groups in consort, a phenomenon known as muscle synergies. Here, we want to analyse if synergies emerge and facilitate learning. 

For \name{} where an agent has to simultaneously learn multiple manipulations / tasks, common patterns emerges and fewer synergies i.e. 12 (Figure \ref{fig:Synergies_iterations}), can explain more than $80\%$ of the variance of the data. Furthermore, we observe that tasks start sharing more synergies (on average 6, see Figure \ref{Fig:CosSimilarity}). This is expected as each task needs a combination of shared (task-aspecific) and  task-specific synergies. Common patterns of activations seems to be related with learning. Indeed, earlier in the training more synergies are needed to explain the same amount of variance of the data. The peak is reached at $12.5k$ iterations where more than $90\%$ of the variance is explained by 12 synergies (see Figure \ref{fig:Synergies_iterations}).

As expected, the expert policies shared fewer common muscle activations as indicated by fewer synergies shared between tasks (on average 2, see Figure \ref{Fig:CosSimilarity}) and by the overall greater number of synergies needed to explain most of the variance: to explain more than $80\%$ of the variance it is needed to use more than 20 synergies. Similar results were obtained with the student policy (on average 1 similar synergies between tasks, see Figure \ref{Fig:CosSimilarity}).

\begin{table}[h]
    \centering
        \begin{tabular}{|l|c|c|}
        \hline
        Object &  Creator & License \\
        \hline
        waterbottle    & badger            & GNU GPL v2\\
        train           & Jason Shoumar     & public domain\\
        airplane        & Gravity Sketch    & CC BY-4.0\\
        wine glass      & Michael Spivey    & CC BY 3.0\\
        cup             & Ablapo            & CC BY 3.0 \\
        mug             & Ryan Smith        & credit, remix, non-commercial \\
        alarm clock     & Javier Ruiz       & CC BY-SA 3.0 \\
        banana          & Lloyd Bolts       & credit, remix, non-commercial \\
        hammer          & Microsoft         & CC BY 4.0 \\
        mouse           & Michael Spivey    & CC BY 3.0 \\
        duck            & willie            & CC0 1.0 \\
        \hline
        \end{tabular}
        \caption{Creators and License for the objects illustrated.}
        \label{Fig:Licenses}
\end{table}

\subsubsection{PreGrasp informed Dexterous Manipulation}
 \label{sec:PGDM}
We adopted Dasari et al \cite{dasari_learning_2023} solution where the desired object trajectory $\hat{X} = [\hat{x}^0, ... , \hat{x}^T ]$ is leveraged to capture the temporal complexities of dexterous manipulation. Additionally hand-object pre-grasp posture $\phi_{object}^{pregrasp}$ is leveraged to guide the search space. For each task, first a trajectory planner is used to solve for the free space movement driving the hand to the pre-shape pose, and then PPO is employed to solve for achieving the desired object trajectory. We extracted relevant pregrasp informations from the GRAB motion capture \cite{taheri_grab_2020-1} dataset which contains high-quality human-object interactions. Note that these details can also be acquired by running standard hand tackers \cite{rong2021frankmocap}
on free form human videos. 

We used the hand posture just before the initial contact with the object (see Figure \ref{fig:Objects}) as the pre-grasp posture. This allows us to not require any physical or geometric information about the object. Given this tracked posture, we recover MyoHand posture via means of  Inverse Kinematics over the finger tips.

\end{document}